\documentclass[10pt,journal,compsoc]{IEEEtran}
\ifCLASSOPTIONcompsoc
  \usepackage[nocompress]{cite}
\else
  \usepackage{cite}
\fi

\usepackage[utf8]{inputenc} 
\usepackage[T1]{fontenc}    
\usepackage[pagebackref,breaklinks,colorlinks]{hyperref}
\usepackage{url}            
\usepackage{booktabs}       
\usepackage{amsfonts}       
\usepackage{nicefrac}       
\usepackage{microtype}      
\usepackage{graphicx}
\usepackage{amsmath}
\usepackage{amssymb}
\usepackage{array}

\usepackage{diagbox}
\usepackage{multicol}
\usepackage{enumerate}
\usepackage{times}
\usepackage{epsfig}
\usepackage{threeparttable}
\usepackage{enumitem}
\usepackage{multirow}
\usepackage{color}
\usepackage{array}
\usepackage{setspace}
\usepackage{makecell}

\newcommand{\review}[1]{\textcolor{black}{#1}}

\begin{document}

\title{Adverse Weather Optical Flow: Cumulative\\Homogeneous-Heterogeneous Adaptation}


\author{Hanyu~Zhou,
        Yi~Chang*,~\IEEEmembership{Member,~IEEE,}
        Zhiwei~Shi,
        Wending~Yan,
        Gang~Chen,~\IEEEmembership{Member,~IEEE,}
        Yonghong~Tian,~\IEEEmembership{Fellow,~IEEE,}
        and~Luxin~Yan,~\IEEEmembership{Member,~IEEE}

        \IEEEcompsocitemizethanks{\IEEEcompsocthanksitem *: Corresponding author.
        \IEEEcompsocthanksitem Hanyu~Zhou, Yi~Chang, Zhiwei~Shi, and Luxin~Yan are with the National Key Laboratory of Multispectral Information Intelligent Processing Technology, School of Artificial Intelligence and Automation, Huazhong University of Science and Technology, Wuhan, China. (e-mail: \{hyzhou, yichang, yanluxin\}@hust.edu.cn)
        \IEEEcompsocthanksitem Wending Yan is with the Huawei International Co. Ltd., Singapore.
        \IEEEcompsocthanksitem Gang Chen is with the School of Computer Science and Engineering, Sun Yat-sen University, Guangzhou, China.
        \IEEEcompsocthanksitem Yonghong Tian is with the School of Electrical Engineering and Computer Science, Peking University, Beijing, China.}}

\IEEEtitleabstractindextext{%
\begin{abstract}
Optical flow has made great progress in clean scenes, while suffers degradation under adverse weather due to the violation of the brightness constancy and gradient continuity assumptions of optical flow. Typically, existing methods mainly adopt domain adaptation to transfer motion knowledge from clean to degraded domain through one-stage adaptation. However, this direct adaptation is ineffective, \review{since there exists a large gap due to adverse weather and scene style between clean and real degraded domains.} Moreover, even within the degraded domain itself, static weather (\emph{e.g.}, fog) and dynamic weather (\emph{e.g.}, rain) have different impacts on optical flow. To address above issues, we explore synthetic degraded domain as an intermediate bridge between clean and real degraded domains, and propose a cumulative homogeneous-heterogeneous adaptation framework for real adverse weather optical flow. Specifically, for clean-degraded transfer, our key insight is that static weather possesses the depth-association homogeneous feature which does not change the intrinsic motion of the scene, while dynamic weather additionally introduces the heterogeneous feature which results in a significant boundary discrepancy in warp errors between clean and degraded domains. For synthetic-real transfer, we figure out that cost volume correlation shares a similar statistical histogram between synthetic and real degraded domains, benefiting to holistically aligning the homogeneous correlation distribution for synthetic-real knowledge distillation. Under this unified framework, the proposed method can progressively and explicitly transfer knowledge from clean scenes to real adverse weather. In addition, we further collect a real adverse weather dataset with manually annotated optical flow labels and perform extensive experiments to verify the superiority of the proposed method. \review{Both the code and the dataset will be available at \emph{https://github.com/hyzhouboy/CH2DA-Flow}.}

\end{abstract}

\begin{IEEEkeywords}
Optical flow, adverse weather, domain adaptation, cumulative transfer.
\end{IEEEkeywords}}

\maketitle
\IEEEdisplaynontitleabstractindextext
\IEEEpeerreviewmaketitle

\IEEEraisesectionheading{\section{Introduction}\label{sec:introduction}}

\begin{figure*}
  \centering
   \includegraphics[width=0.99\linewidth]{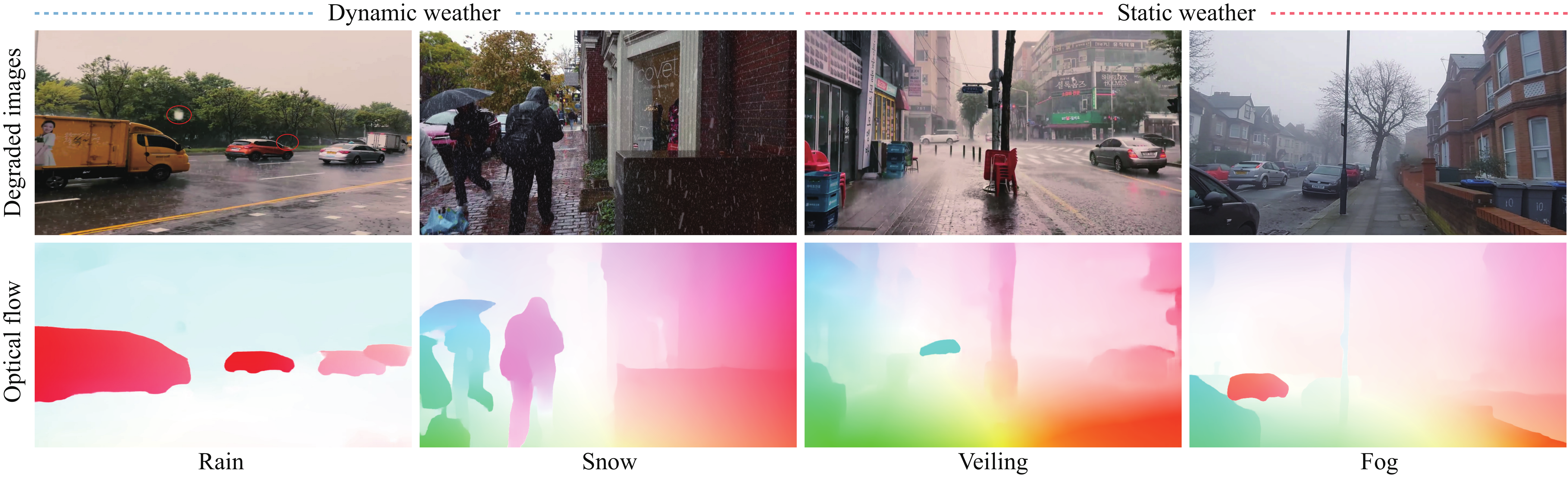}
   \caption{Optical flow visualization of the proposed method CH$^2$DA-Flow under adverse weather. The proposed method can well estimate the high-quality optical flows in various degraded conditions, including dynamic weather (\emph{e.g.}, rain and snow) and static weather (\emph{e.g.}, veiling and fog).
   }
   \label{Adverse_Weather}

\end{figure*}

\IEEEPARstart{O}{ptical} flow plays an important role in scene motion perception for autonomous driving \cite{menze2015object}. Optical flow aims to model the temporal correspondence between adjacent frames for the intrinsic motion of the scene, which has been widely applied in different downstream vision tasks, such as video object detection \cite{cui2021tf} and tracking \cite{fan2019lasot}. Learning-based optical flow methods are divided into two categories: supervised and unsupervised. Supervised methods \cite{dosovitskiy2015flownet, sun2018pwc, hui2018liteflownet, teed2020raft} rely heavily on optical flow labels, but it is difficult to obtain accurate optical flow labels in real scenes. Unsupervised methods \cite{jason2016back, ren2017unsupervised, meister2018unflow, liu2019selflow} alleviate the dependency on motion labels, and resort to the brightness constancy and gradient continuity assumptions to learn optical flow. However, these methods have made great progress in clean scenes, but suffer challenges under adverse weather. This is because degradation breaks the basic assumptions which optical flow depends on. \review{Our goal is to estimate optical flow for the intrinsic scene motion under various adverse weather conditions, as shown in Fig. \ref{Adverse_Weather}}.

Adverse weather is divided into dynamic weather (\emph{e.g.}, rain and snow) and static weather (\emph{e.g.}, fog and veiling). Dynamic weather with changing luminance on the imaging plane breaks the brightness constancy assumption of optical flow, as an independent motion mixed with the intrinsic motion of the scene, thus exacerbating the difficulty of optical flow estimation. Static weather that weakens image contrast breaks the gradient continuity assumption of optical flow and degrades the discriminative visual feature, thus causing the motion feature mismatching and limiting the performance of optical flow. \review{Dynamic weather and static weather correspond to optical occlusion and scattering effects in the physical space, linear occlusion and multiplicative attenuation in the visual space, and continuous disturbance and stationary effects in the motion space. In other words, the two weather conditions affect optical flow differently in the physical, visual, and motion spaces.}

\begin{figure*}
  \centering
   \includegraphics[width=1.00\linewidth]{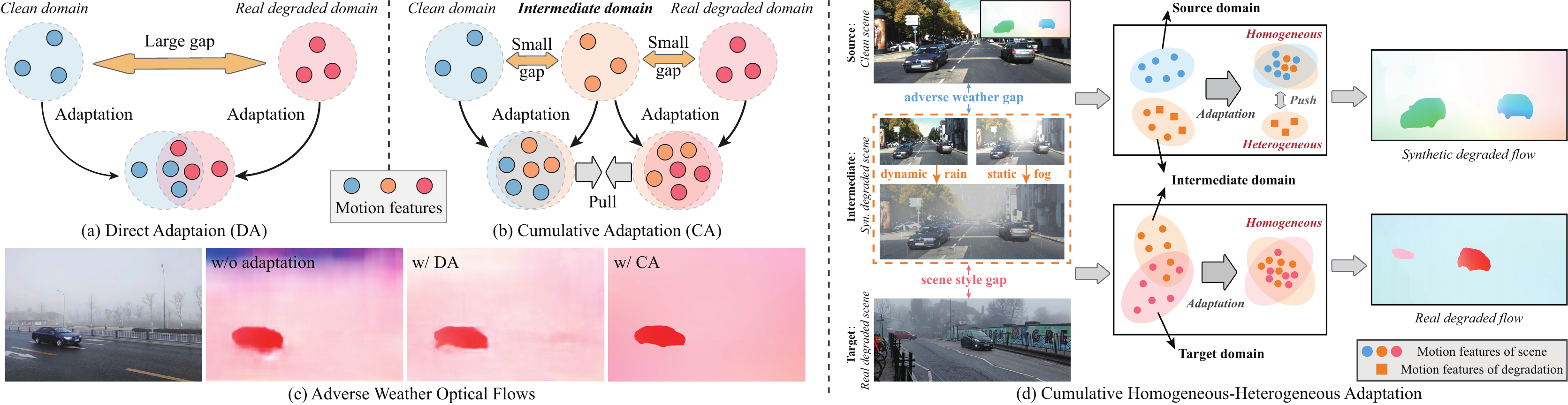}
   \caption{Illustration of adaptation paradigms for adverse weather optical flow. (a) Direct adaptation. (b) Cumulative adaptation. (c) Adverse weather optical flows. (d) Cumulative homogeneous-heterogeneous adaptation. \review{Direct adaptation mainly transfers motion knowledge from clean to degraded domain via one-stage adaptation.} \review{However, this direct adaptation usually suffers from domain shift due to the large domain gap between clean and real degraded domains. To address this issue, we propose a cumulative adaptation framework, which introduces an intermediate domain to close the large gap, making the motion features of the intermediate domain aligned with the features of the two domains, respectively.
   Considering that dynamic weather (\emph{e.g.}, rain) and static weather (\emph{e.g.}, fog) have different impacts on optical flow even within the degraded domain, we further extend the cumulative adaptation framework into the cumulative homogeneous-heterogeneous adaptation framework, which not only aligns the cross-domain homogeneous motion features of the scene, but also strips away the heterogeneous motion features of the dynamic degradation.}
   }
   \label{Related_Work}

\end{figure*}

There are two categories of methods for adverse weather optical flow: image restoration in visual space and knowledge transfer in motion space. Image restoration can improve the visual effect of the image sequence in the spatial dimension, but it cannot guarantee the accurate motion matching in the temporal dimension. In contrast, \review{knowledge transfer circumvents the challenges of directly estimating optical flow from degraded or restored images by formulating adverse weather optical flow as a task of domain adaptation, which transfers motion knowledge from clean domain to degraded domain through one-stage adaptation, as shown in Fig. \ref{Related_Work} (a)}.
For example, Li \emph{et al.} \cite{li2018robust, li2019rainflow} directly learned the motion knowledge mapping from clean scene to rainy scene in image level \cite{li2018robust} and feature level \cite{li2019rainflow}, respectively. Yan \emph{et al.} \cite{yan2020optical} proposed the motion consistency to model the knowledge transfer between the paired clean and synthetic foggy images. However, \review{this direct adaptation usually suffers from inter-domain gap and intra-domain discrepancy.} \review{As for the inter-domain gap, there exists a large domain gap caused by adverse weather and scene style between clean and real degraded domains, exacerbating the feature distribution discrepancy between the two domains.} \review{As for the intra-domain discrepancy, even within the degraded domain itself, there exist obvious differences between the degradation patterns of static weather and dynamic weather on optical flow.} The two issues limit the performance of this direct adaptation under real adverse weather in Fig. \ref{Related_Work} (c). To overcome the above problems, we introduce synthetic degraded domain as an intermediate bridge between clean and real degraded domains to reinforce the feature alignment in Fig. \ref{Related_Work} (b), and propose a novel cumulative adaptation framework for adverse weather optical flow in Fig. \ref{Related_Work} (d), which \review{\emph{explores the cross-domain homogeneous and heterogeneous features}.}

As for the clean-degraded gap, clean and degraded images are the same scene. First, we discover that static weather possesses \review{the depth association with the intrinsic scene motion, which is the homogeneous feature that cannot be changed due to static weather.} \review{Regarding depth association,} on the one hand, static weather affects image contrast along depth through atmospheric scattering model \cite{narasimhan2002vision}, thus affecting the smoothness of the intrinsic scene motion along the depth; on the other hand, there exists a natural 2D-3D geometry projection relationship \cite{zhou2017unsupervised} between depth and optical flow, revealing that depth can in turn enhance the intrinsic scene motion. Second, \review{we theoretically analyze that dynamic weather introduces additional disturbances into the scene motion, causing a significant boundary discrepancy between clean and degraded domains. This boundary discrepancy is the heterogeneous feature caused by dynamic weather represented in the warp error, which denotes the residual image between the aligned adjacent images.} Therefore, we choose the depth association to consistently guide the directional homogeneous intrinsic motion knowledge transfer, and the warp error to contrastively model the heterogeneous boundary knowledge for inhibiting dynamic interference.

As for the synthetic-real gap, since synthetic degraded and real degraded domains usually exist in various scenes, the discrepancy between them is not caused by adverse weather but by scene style, which is the inherent visual characteristic of the image. To this end, we mainly focus on exploiting the homogeneous feature related to motion rather than the heterogeneous feature caused by scene style. \review{Note that the cost volume is a typical motion feature, which can physically measure the similarity between adjacent frames to describe the global motion trend, regardless of synthetic and real degraded images}. We figure out that synthetic and real degraded domains share a similar statistical histogram of cost volume correlation, which represents the holistic motion distribution. This drives us to holistically align the homogeneous correlation distribution for explicitly transferring motion knowledge from synthetic degraded to real degraded domain.

In this work, we propose a novel cumulative homogeneous-heterogeneous domain adaptation (CH$^2$DA-Flow) framework for real adverse weather optical flow, including clean-degraded motion adaptation (CDMA) and synthetic-real motion adaptation (SRMA). During the CDMA stage, we design a depth association homogeneous motion adaptation for static weather and a warp error heterogeneous boundary adaptation for dynamic weather. In the depth association homogeneous motion adaptation, \review{we utilize depth estimated from stereo clean images to bridge the optical flow of clean domain and the image of degraded domain using physical relationship, thus transferring homogeneous intrinsic motion knowledge from clean to degraded domain via motion consistency.}
In the warp error heterogeneous boundary adaptation, \review{We use the estimated optical flow to align the adjacent clean images and the adjacent synthetic degraded images via warp operation, respectively, and then obtain the corresponding warp error by subtraction,} where we further distinguish the heterogeneous boundary errors caused by dynamic disturbance to push away using contrastive learning \cite{park2020contrastive}. During the SRMA stage, we first transform the synthetic and real degraded images to the cost volume space. Then, we use K-L divergence \cite{kullback1951information} to measure the holistic distance of homogeneous correlation values between the two domains, and constrain the holistic motion distribution by minimizing this distance function, thus distilling the motion knowledge from synthetic degraded domain to real degraded domain. \review{Under this unified training framework, the cumulative framework promises the overall stability of motion knowledge transfer, and the homogeneous-heterogeneous adaptation guarantees the robustness of the final optical flow model to various adverse weather conditions.}

This work is an extension of our earlier publication in CVPR 2023 \cite{zhou2023unsupervised}, with four main improvements on \review{problem}, adaptation method, dataset, and experiment. First, different from the previous version which only handles the optical flow under static weather, \review{we further introduce the warp error to learn the impact pattern of dynamic weather on optical flow, thus improving the optical flow under various adverse weather conditions}. Second, we incorporate the homogeneous-heterogeneous adaptation technique into the cumulative adaptation framework from the previous version. The previous version proposes a cumulative framework for progressive knowledge transfer but is limited in representing the discriminative motion feature for adverse weather. To this end, we further focus on \review{designing a homogeneous-heterogeneous motion adaptation between clean and degraded domains.} Third, since existing adverse weather motion datasets contain relatively single scenes and few samples, we propose a real adverse weather dataset with manually annotated optical flow labels, which covers multiple moving objects and scenarios under rain, fog, and snow. Fourth, \review{we conduct more quantitative and qualitative experiments to verify the superiority of the proposed method under various adverse weather conditions}. Overall, our main contributions are summarized as follows:
\begin{itemize}[leftmargin=10pt]
\item We propose a cumulative homogeneous-heterogeneous adaptation framework for adverse weather optical flow. \review{The cumulative framework can promise the overall stability of motion knowledge transfer from clean to real degraded domain, and the homogeneous-heterogeneous adaptation technique can ensure the knowledge transfer to various adverse weather conditions}.

\item \review{We discover that static weather possesses the depth association with the intrinsic scene motion which does not change, while dynamic weather causes a significant boundary discrepancy in warp errors between clean and degraded domains. This drives us to design a homogeneous motion adaptation for static weather and a heterogeneous boundary adaptation for dynamic weather, achieving clean-degraded knowledge transfer.}

\item We illustrate that synthetic and real degraded domains share a similar correlation statistical histogram, motivating us to holistically align the homogeneous correlation distribution for synthetic-real knowledge distillation.

\item We build a real adverse weather dataset with optical flow labels, and conduct extensive experiments to verify the superiority and generalization of the proposed method for different adverse weather conditions, such as rain, fog, and snow.
\end{itemize}

\section{Related Work}
\label{sec:relatedwork}
\textbf{Optical Flow.}
Optical flow is a task of estimating per-pixel temporal correspondence between video frames for the intrinsic motion of the scene.
Traditional methods \cite{sun2010secrets, revaud2015epicflow, kroeger2016fast, hur2017mirrorflow} often formulate optical flow as an energy minimization problem, while suffering from restricted limitations in computational efficiency and performance. In recent years, learning-based optical flow methods \cite{dosovitskiy2015flownet, jason2016back, ilg2017flownet, ranjan2017optical, lai2017semi, ren2017unsupervised, sun2018pwc, hui2018liteflownet, liu2020learning, luo2021upflow, chi2021feature, zhang2021separable, jiang2021learning, aleotti2021learning, sun2021autoflow, huang2022flowformer} have been proposed to improve motion feature representation, including supervised and unsupervised approaches. Supervised methods \cite{luo2022learning, sun2018pwc, ilg2017flownet, hur2019iterative, teed2020raft, jiang2021learning, Luo_2023_ICCV, Shi_2023_CVPR, Dong_2023_CVPR} mainly construct various deep network structures to learn motion field using optical flow labels. For example, Dosovitskiy \emph{et al.} \cite{dosovitskiy2015flownet} proposed the first end-to-end supervised optical flow network on synthetic dataset.
Sun \emph{et al.} \cite{sun2018pwc} designed an efficient network PWC-Net by warping images with elegant cost volume in a coarse-to-fine pyramid. RAFT \cite{teed2020raft} was an important development of PWC-Net and constructed 4D cost volume for all pairs of pixels. To further improve motion feature presentation, transformer was incorporated into optical flow architecture, \emph{e.g.}, GMA \cite{jiang2021transformer} and TransFlow \cite{lu2023transflow}, which achieved better performance than RAFT. However, due to the difficulty in obtaining motion ground truth in real world, these supervised methods may have limited performance in real world. Furthermore, to relieve the dependency on motion labels, unsupervised methods \cite{jonschkowski2020matters, liu2019ddflow,liu2019selflow, meister2018unflow, ren2017unsupervised, yang2019volumetric, jason2016back, zhao2020maskflownet, zhou2017unsupervised, wang2018occlusion, ranjan2019competitive, liu2020learning, chi2021feature, Yuan_2023_ICCV} achieve general attention. Unsupervised methods follow the basic architecture of supervised methods \cite{sun2018pwc,teed2020raft}, but the main difference is that the training process requires the brightness constancy and gradient continuity assumptions of optical flow. Although they have achieved satisfactory results in clean scenes, they would suffer degradation under adverse weather. The main reason is that adverse weather breaks the basic assumption of optical flow. In this work, we aim to deal with optical flow under adverse weather in Fig. \ref{Adverse_Weather}, including dynamic weather (\emph{e.g.}, rain and snow) and static weather (\emph{e.g.}, veiling and fog).

\begin{figure*}
  \centering
   \includegraphics[width=1.0\linewidth]{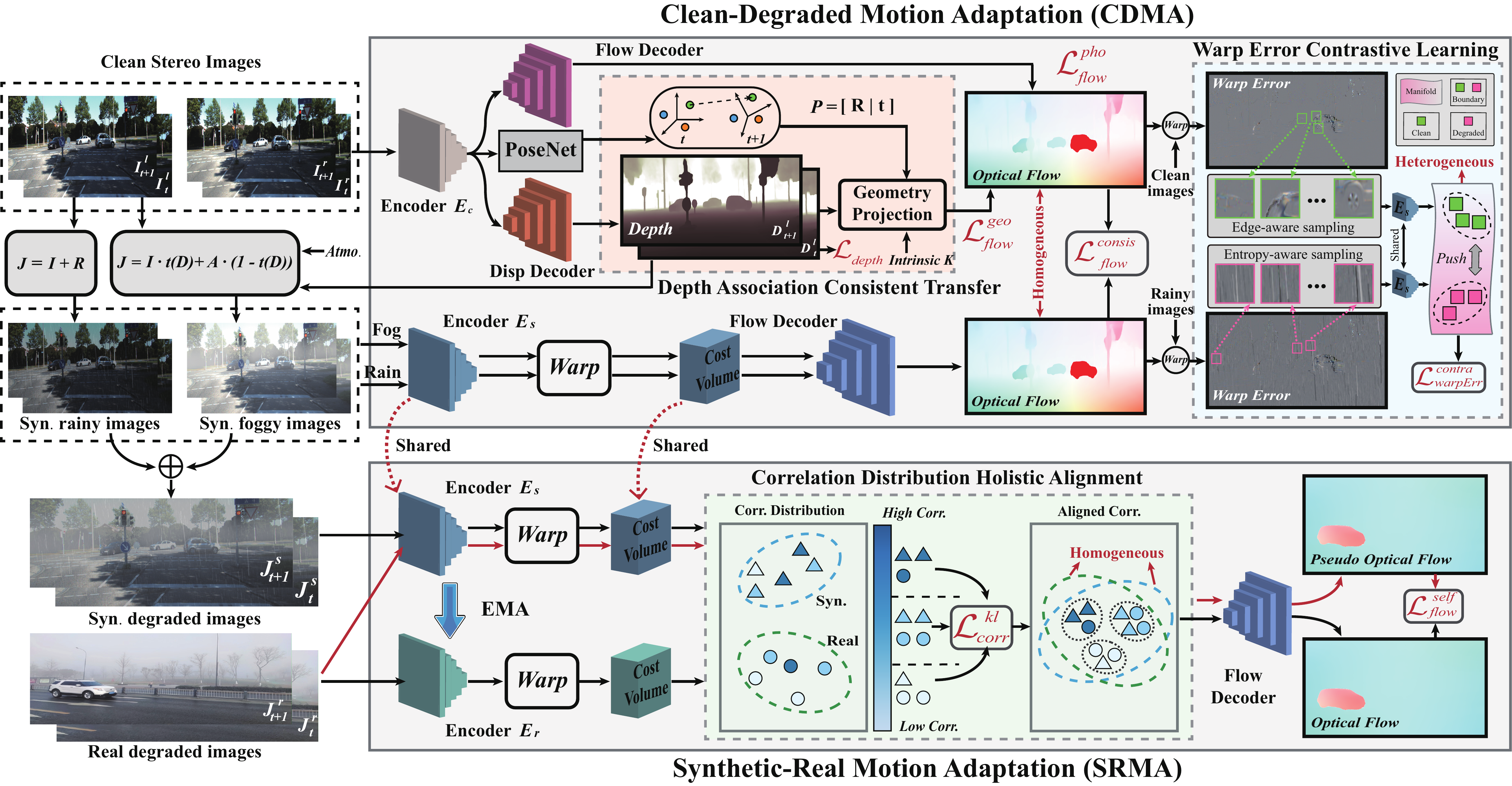}
   \caption{The architecture of the proposed CH$^2$DA-Flow method mainly contains clean-degraded motion adaptation (CDMA) and synthetic-real motion adaptation (SRMA). During the CDMA stage, we design the depth association homogeneous motion adaptation for static weather, and the warp error heterogeneous boundary adaptation for dynamic weather, thus directionally transferring motion knowledge from the clean domain to the synthetic degraded domain. During the SRMA stage, we propose the cost volume homogeneous correlation adaptation, which builds the correlation distribution holistic alignment module to distill motion knowledge of the synthetic degraded domain to the real degraded domain.
   }
   \label{Framework}

\end{figure*}

\noindent
\textbf{Adverse Weather Optical Flow.}
There are two typical approaches: image restoration and knowledge transfer. Image restoration is to perform the image deraining \cite{fu2017removing, liu2021unpaired, yan2021self, yang2017deep, zhang2018density, ye2021closing, li2019heavy} or defogging \cite{liu2019griddehazenet, qin2020ffa, ren2018gated, shao2020domain, wu2021contrastive} with subsequent optical flow estimation. However, image restoration does improve the spatial visual quality of image sequence, but it fails to ensure the temporal motion matching due to the over-smooth details or residual degradation. Different from image restoration, knowledge transfer formulates this problem as a domain adaptation task, which transfers motion knowledge from clean to degraded domain through one-stage adaptation.
For example, Li \emph{et al.} designed a rain-invariant prior to directly model the motion knowledge mapping from clean scene to rainy scene in image level \cite{li2018robust} and feature level \cite{li2019rainflow}. Li \emph{et al.} \cite{li2021gyroflow} resorted to gyroscope that is robust to environment to model the clean motion of background, which is then transferred to the optical flow of the degraded domain, while failed to model the motion of the independent moving object. Yan \emph{et al.} \cite{yan2020optical} proposed the motion consistency to model the knowledge transfer between the paired clean and synthetic foggy images.
Zhou \emph{et al.} \cite{Zhou_aaai} further modeled the motion discrepancy between clean and degraded domains to improve the representation of degraded optical flow.
Schmalfuss \emph{et al} \cite{Schmalfuss_2023_ICCV} utilized an adversarial attack strategy \cite{yuan2019adversarial} to synthesize degradation on clean images, and learned the motion mapping between clean and degraded domains. \review{However, this direct adaptation neglects the large inter-domain gap due to adverse weather and scene style between clean and real degraded domains in Fig. \ref{Related_Work} (a)}. Moreover, even within the degraded domain itself, dynamic weather and static weather affect optical flow differently, thus limiting the optical flow performance under adverse weather. In this work, we propose a novel cumulative adaptation framework in Fig. \ref{Related_Work} (b), \review{which introduces synthetic degraded domain as an intermediate bridge between clean domain and real degraded domain to reinforce the feature alignment.}

\noindent
\textbf{Adverse Weather Scene Understanding.}
Knowledge transfer has been generally applied for different adverse weather scene understanding tasks, such as, semantic segmentation \cite{ma2022both, sakaridis2018semantic, sakaridis2018model, liu2021learning}, object detection \cite{xu2020cross, tian2021knowledge} and depth estimation \cite{saunders2023self, gasperini2023robust}. For example, Dai \emph{et al.} \cite{dai2020curriculum} took a weight sharing strategy to directly transfer semantic knowledge from clean scene to foggy scene. Tian \emph{et al.} \cite{tian2021knowledge} designed a shared classifier to align the common attribute of objects between clean and foggy images. Saunders \emph{et al.} \cite{saunders2023self} transformed clean images to adverse weather domain, and proposed to learn the depth model corresponding to the degraded images using a pseudo-supervised strategy. However, the knowledge transfer process of these methods is too implicit due to the lack of intrinsic characteristic knowledge of the target vision task, resulting in limited model performance. \review{In this work, we explore the explicit features between clean and degraded domains,} as a guidance to transfer motion knowledge from clean domain to degraded domain, thus further improving the optical flow performance under adverse weather.

\noindent
\textbf{Domain Adaptation.}
Domain adaptation \cite{ben2006analysis, wang2018deep, chen2022msdn, ganin2015unsupervised, you2019universal, tzeng2017adversarial, chen2022deliberated} is to solve the feature discrepancy between source domain and target domain. \review{Domain adaptation can be divided into two categories: homogeneous adaptation \cite{chen2021hsva, long2018conditional, saito2018maximum, lee2019sliced} and heterogeneous adaptation \cite{tan2015transitive, tan2017distant, Zhou_aaai}. Homogeneous adaptation aims to align the similar features between source and target domains, which are named homogeneous features. Heterogeneous adaptation focuses on peeling off the features with discrepancy between source and target domains, which are named heterogeneous features.}
Degraded scene understanding \cite{zhou2023unsupervised, gao2022cross, sakaridis2018semantic, sakaridis2019guided} can be formulated as a domain adaptation task, which focuses on transferring specific knowledge from source clean domain to target degraded domain. \review{Most of the existing methods \cite{zhou2023unsupervised, gao2022cross, ma2022both} adopt the homogeneous adaptation to transfer specific knowledge from clean to degraded domain.} However, in degraded scenes, the heterogeneous feature caused by degradation interference would diverge the entire feature distribution, \review{resulting in that the homogeneous adaptation has the potential risk of misaligning the heterogeneous features into the homogeneous feature space}. To this end, we propose a homogeneous-heterogeneous adaptation technique for degraded optical flow.

\section{Cumulative Hetero-Homogeneous Adaptation for Adverse Weather Optical Flow}
\subsection{Overall Framework}
Adverse weather optical flow is formulated as a task of knowledge transfer from source clean domain to target real degraded domain. \review{However, there exist a large inter-domain gap between clean and real degraded domains, and a large intra-domain discrepancy within the degraded domain, thus limiting the optical flow performance under real adverse weather.} In this work, we introduce synthetic degraded domain as an intermediate bridge between clean and real degraded domains, and propose a novel cumulative homogeneous-heterogeneous adaptation framework for adverse weather optical flow in Fig. \ref{Framework}, which not only closes the inter-domain gap but also alleviates the intra-domain discrepancy. \review{The whole framework structure looks complicated but is simply modularized into two modules: pixel-aligned clean-degraded motion adaptation and pixel-misaligned synthetic-real motion adaptation.} As for pixel-aligned clean-degraded motion adaptation, we design depth association homogeneous motion adaptation for static weather and warp error heterogeneous boundary adaptation for dynamic weather. In depth association homogeneous motion adaptation, we utilize depth to associate the degradation and optical flow with physical relationship, thus consistently transferring global motion knowledge from clean to degraded domain. In warp error heterogeneous boundary adaptation, we introduce warp error to model the boundary discrepancy between clean and degraded domains, thus contrastively transferring the local boundary knowledge to degraded domain and inhibiting additional dynamic disturbance. As for pixel-misaligned synthetic-real motion adaptation, we design cost volume homogeneous correlation adaptation, which explores the similar correlation distribution in the cost volume space to transfer motion knowledge from synthetic to real domain. Under this unified framework, the proposed cumulative homogeneous-heterogeneous adaptation can progressively and explicitly transfer motion knowledge from clean scene to real adverse weather. Next, we will describe the proposed cumulative adaptation framework from two perspectives: what to transfer and how to transfer.

\subsection{Mechanism of Degraded Optical Flow}
\label{sec:Mechanism}
\begin{figure}
  \centering
   \includegraphics[width=0.99\linewidth]{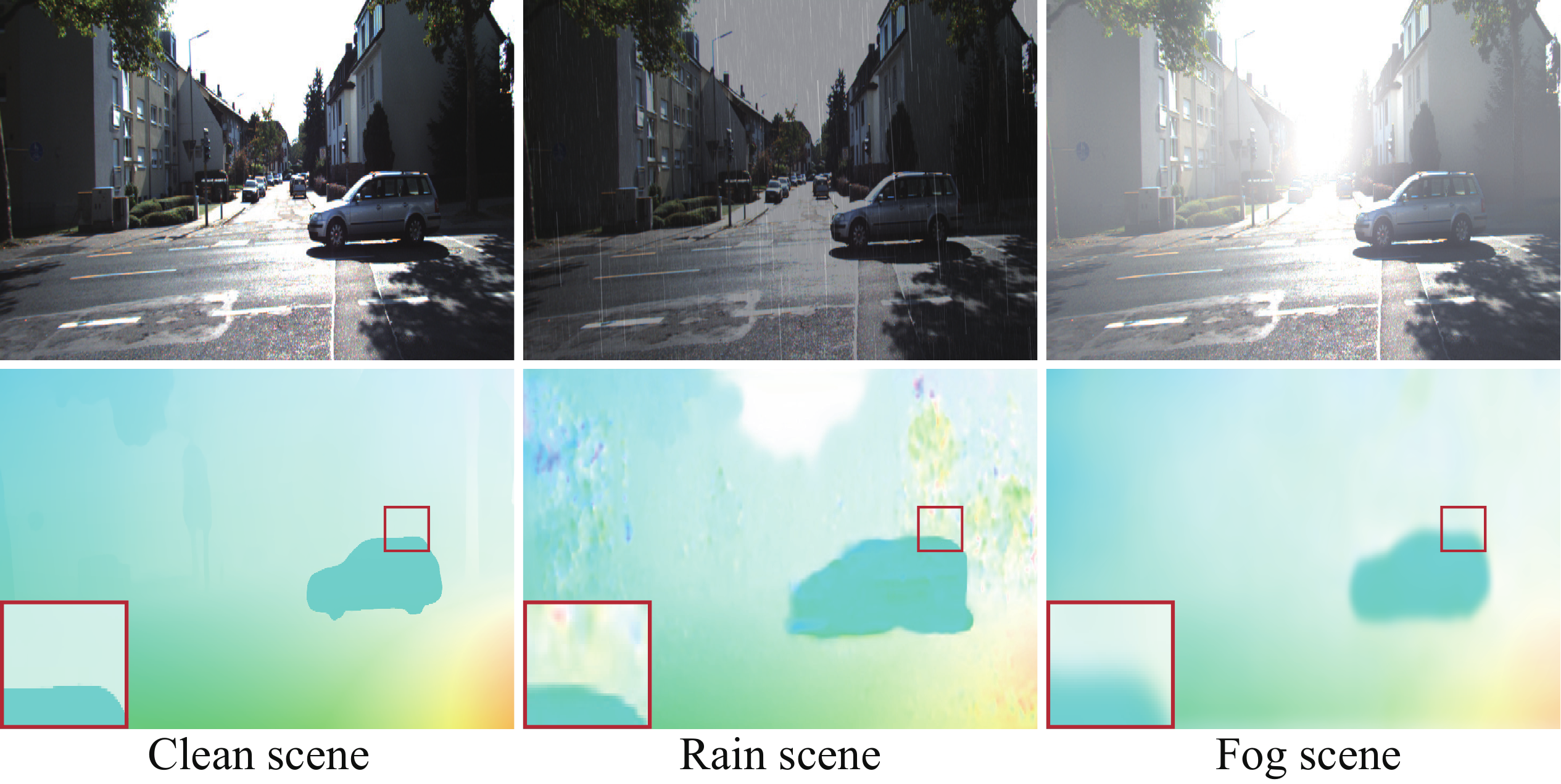}
   \caption{Visualization of optical flow under various weather conditions. Compared with the optical flow of clean scene, rain brings many additional artifacts in the optical flow while fog smoothens the entire optical flow. This motivates us to categorize the impact of adverse weather into dynamic weather and static weather for different motion adaptation strategies.
   }
   \label{Degradation}

\end{figure}

There are different adverse weather conditions in real scenes, such as, rain, snow, fog and veiling. These degradations are mainly divided into two categories: dynamic weather and static weather. Under adverse weather, dynamic weather with rapidly falling speed appears in the camera, while static weather that remains relatively stationary with the background affects the image contrast along scene depth. Scene content is transmitted to the camera imaging plane through optical occlusion of dynamic weather and scattering effect of static weather in the physical space. The degradation mechanism \cite{li2020all} to image content is as follows:
\begin{equation}
	\setlength\abovedisplayskip{4pt}
  \setlength\belowdisplayskip{4pt}
\begin{aligned}
  J = t \cdot (I + R) + (1 - t) \cdot A,
  \label{eqa:degradation_mechanism}
\end{aligned}
\end{equation}
where $J$, $I$ is the degraded image and clean image, respectively. $A$ is the atmospheric light, $t$ denotes the decay function related to static weather and $R$ denotes the dynamic weather component. Eq. \ref{eqa:degradation_mechanism} reveals that dynamic weather and static weather are essentially linear degradation and multiplicative degradation, respectively. In order to further analyze the impact of different degradations on optical flow, we use rain and fog as the degradation examples to visualize the estimated optical flow results corresponding to clean images, synthetic rainy images and synthetic foggy images in the same scene in Fig. \ref{Degradation}.
We have two observations. First, rain streaks show changing luminance, producing linear occlusion in the visual space, and breaking the brightness constancy assumption of optical flow, thus playing as an independent motion mixed with the intrinsic motion of the scene in the motion space. Second, fog weakens image contrast, and multiplicatively attenuates the image texture in the visual space which violates the gradient continuity assumption of optical flow, thus oversmoothing the intrinsic motion of the scene in the motion space. In other words, dynamic weather and static weather have different influences on optical flow in the physical space, visual space and motion space, making it difficult to use the unified domain adaptation to handle optical flow estimation in multiple weather conditions. \review{In this work, we model the homogeneous and heterogeneous characteristics under dynamic weather and static weather, respectively, serving as a guidance to transfer motion knowledge from clean scene to adverse weather.}

\subsection{Clean-Degraded Motion Adaptation}
As analyzed in Sec. \ref{sec:Mechanism}, static weather and dynamic weather have different impacts on optical flow. In this section, we will describe how to design the homogeneous and heterogeneous adaptation techniques for the two weather conditions in detail.

\subsubsection{Depth Association Homogeneous Adaptation}
Estimating optical flow in static weather scenes is a difficult matter since degradation breaks the optical flow basic assumption. We bypass the difficulty in directly estimating optical flow from degraded images, and formulate the static weather scene optical flow as a domain adaptation task, \review{which focuses on exploiting the homogeneous feature between clean and degraded domains, and designing a reasonable consistent transfer strategy.}

\begin{figure*}
  \centering
   \includegraphics[width=0.99\linewidth]{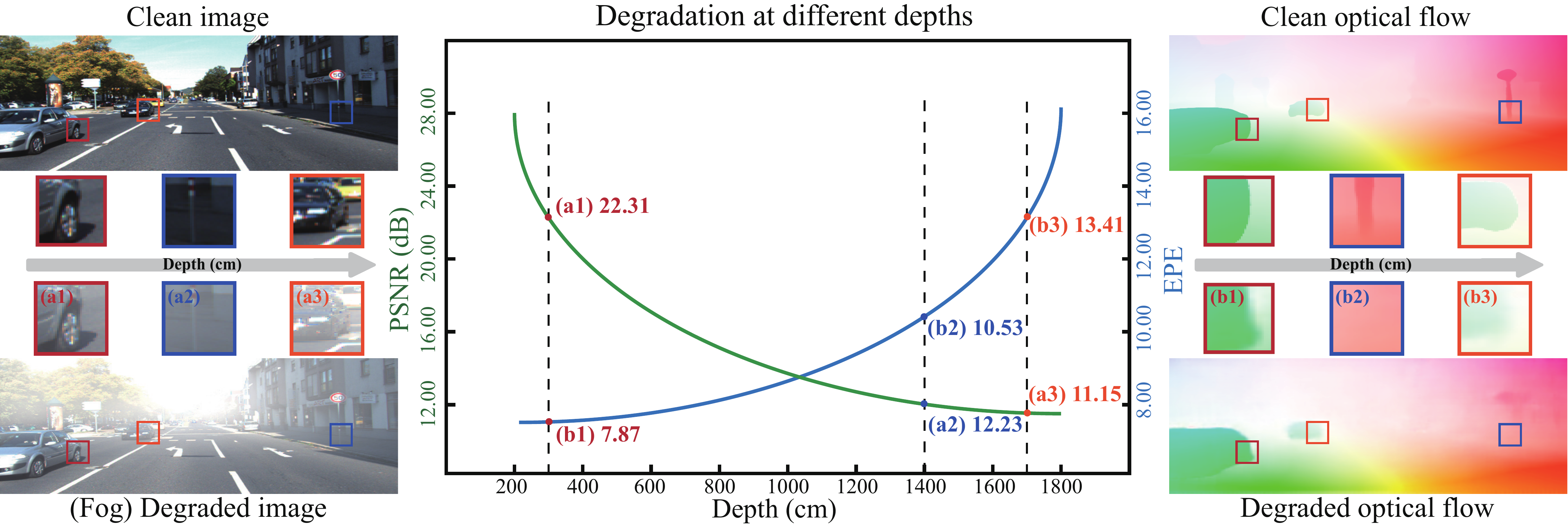}
   \caption{Analysis of static weather (\emph{e.g.}, fog) at different depths. The deeper the depth, the more severe the image, and the more degraded the optical flow. Depth is the key to static weather, motivating us to build a depth association strategy for bridging the clean-degraded knowledge transfer.
   }
   \label{DepthAnalysis}

\end{figure*}

\noindent
\textbf{\review{Homogeneous Intrinsic Motion.}}
In Fig. \ref{Degradation}, different from dynamic weather, static weather cannot bring in additional independent degraded motion. This makes us consider that degradation does not change the intrinsic motion of the scene under static weather, serving as an homogeneous knowledge to transfer. \review{As we know, static weather is a non-uniform degradation related to depth. This prompts us to naturally consider whether the optical flow affected by static weather might also be related to depth.}
To illustrate this, we take clean KITTI2015 \cite{menze2015object} and the corresponding synthetic Fog-KITTI2015 as the experiment datasets, and conduct an analysis experiment on the influence of fog on the image and optical flow along different depths in Fig. \ref{DepthAnalysis}. Compared to the corresponding clean images, we count the PSNR and the optical flow EPE of the foggy images at different depths. As the depth value becomes larger, the lower the PSNR, the higher the optical flow EPE, which means that the degradation of the image and optical flow aggravates with the larger depth. Moreover, we visualize the images (Fig. \ref{DepthAnalysis} (a1)-(a3)) and optical flows (Fig. \ref{DepthAnalysis} (b1)-(b3)) at three depths. We can observe, as the depth increases, the image contrast decreases and the optical flow boundary becomes blurrier, indicating that depth is the key to bridging the clean domain and the static degraded domain. On one hand, depth is associated with static weather (\emph{e.g.}, fog) through atmospheric scattering model \cite{narasimhan2002vision}. On the other hand, depth obeys the strict geometric projection equation \cite{zou2018df} with the rigid part of optical flow. \review{Therefore, there exists a natural physical relationship between static weather, depth and optical flow. This relationship can be used as a physical constraint to associate the intrinsic motion and degraded image, guiding the knowledge transfer from clean to synthetic degraded domain.}

\begin{figure*}
  \centering
   \includegraphics[width=0.99\linewidth]{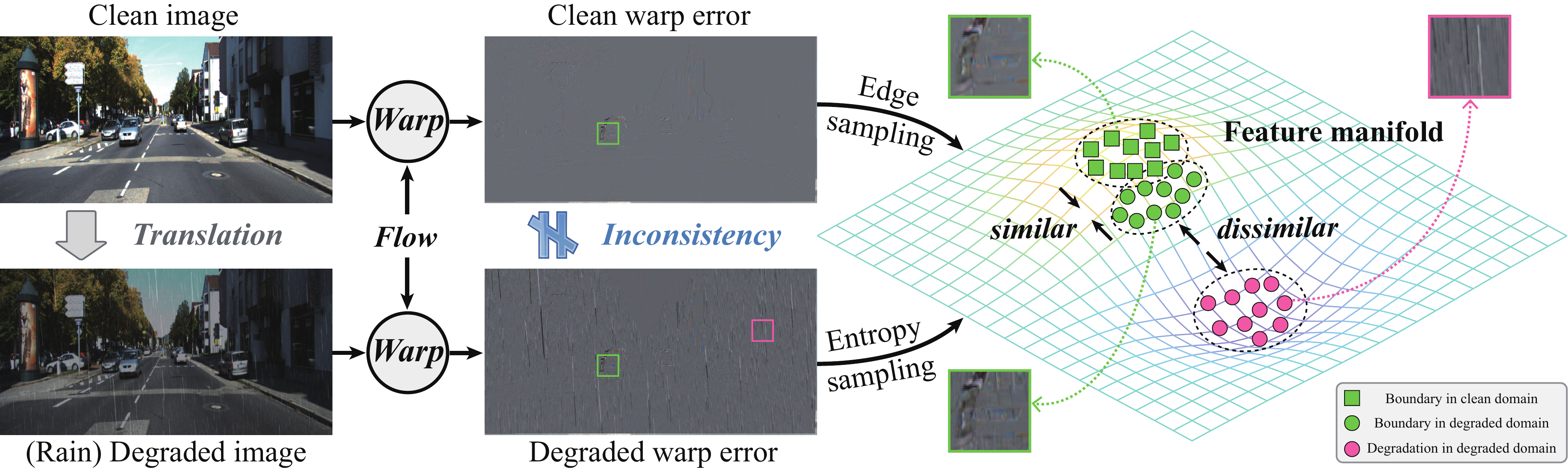}
   \caption{Feature distributions of warp errors between clean domain and (dynamic rain) degraded domain. We encode the clean warp error patches sampled by edge-aware sampling and the degraded warp error patches sampled by entropy-aware sampling into the same feature manifold, where the boundary error features between the two domains are similar while the degraded error features are dissimilar from the boundary error features. This drives us to build contrastive learning for modeling the boundary discrepancy in warp errors between clean and degraded domains.
   }
   \label{WarpError}

\end{figure*}

\noindent
\textbf{Depth Association Consistent Transfer.}
\review{Before motion knowledge transfer from clean to degraded domain, we first enable the optical flow model of the clean domain to learn the capability of estimating the scene motion. Given clean adjacent frames $[I_t^l, I_{t+1}^l]$, we take FlowFormer \cite{huang2022flowformer} as the optical flow backbone to estimate the optical flow $F_c$ of the clean domain with photometric loss \cite{jason2016back}:}
\begin{equation}
  \setlength\abovedisplayskip{4pt}
  \setlength\belowdisplayskip{4pt}
\begin{aligned}
  \mathcal{L}^{pho}_{flow} &=  \resizebox{0.71\hsize}{!}{$\sum\nolimits{\psi(I_t^l - warp(I_{t+1}^l))}\odot(1-O_f)/\sum\nolimits{(1-O_f)}$} \\
   &+ \resizebox{0.70\hsize}{!}{$\sum\nolimits{\psi(I_{t+1}^l - warp(I_t^l))}\odot(1-O_b)/\sum\nolimits{(1-O_b)}$},
  \label{eq:photometric_loss}
\end{aligned}
\end{equation}
\review{where $warp$ is the warping operator, $\psi$ is a sparse $L_p$ norm ($p = 0.4$). $O_f$ and $O_b$ are the forward and backward occlusion masks by checking forward-backward consistency \cite{zou2018df}, and $\odot$ is a matrix element-wise multiplication.} Next, we add right clean adjacent frames $[I_t^r, I_{t+1}^r]$ to left clean adjacent frames $[I_t^l, I_{t+1}^l]$ used for optical flow to form a pair of stereo adjacent frames for depth estimation. In the same way as optical flow, the depths $[D_t^l, D_{t+1}^l]$ are obtained via DispNet \cite{xu2020aanet} using photometric loss:
\begin{equation}
	\setlength\abovedisplayskip{4pt}
  \setlength\belowdisplayskip{4pt}
\begin{aligned}
  \mathcal{L}_{depth} &=  \resizebox{0.60\hsize}{!}{$\sum\nolimits{\psi(I_t^l - warp(I_t^r))}+|{\triangledown}^2D_t^l|e^{-|{\triangledown}^2I_t^l|}$} \\
   &+ \resizebox{0.74\hsize}{!}{$\sum\nolimits{\psi(I_{t+1}^l - warp(I_{t+1}^r))}+|{\triangledown}^2D_{t+1}^l|e^{-|{\triangledown}^2I_{t+1}^l|}$}.
  \label{eqa:depth_loss}
\end{aligned}
\end{equation}

Here depth is used to physically associate with the intrinsic optical flow of the scene and degradation factor simultaneously. For the relationship between depth and intrinsic optical flow, we wish to establish the dense pixel correspondence between the two adjacent frames through depth. Let $p_t$ denote the 2D homogeneous coordinate of a pixel in frame $I_t^l$ and $K$ denote the camera intrinsic matrix. We can compute the corresponding point of $p_t$ in frame $I_{t+1}^l$ using the geometry projection equation \cite{zou2018df} as follows:
\begin{equation}
	\setlength\abovedisplayskip{4pt}
  \setlength\belowdisplayskip{4pt}
\begin{aligned}
  p_{t+1} = KPD_t^l(p_t)K^{-1}p_t,
  \label{eqa:geometry_equation}
\end{aligned}
\end{equation}
where $P$ is the relative camera motion estimated by the pre-trained PoseNet \cite{kendall2015posenet}. We can then compute the rigid flow $F_c^{rigid}$ of clean domain at pixel $p_t$ in $I_t^l$ as follows:
\begin{equation}
	\setlength\abovedisplayskip{4pt}
  \setlength\belowdisplayskip{4pt}
\begin{aligned}
  F_c^{ri}(p_t) = p_{t+1} - p_t.
  \label{eqa:rigid_flow}
\end{aligned}
\end{equation}

To associate depth with optical flow, we further enhance the rigid region of the intrinsic motion of the scene with the consistency prior between the geometrically computed rigid flow and the directly estimated optical flow as follows:
\begin{equation}
\begin{aligned}
  \mathcal{L}_{flow}^{geo} = \sum\nolimits ||F_c - F_c^{ri}||_1 \odot (1-V) / \sum\nolimits (1 - V),
  \label{eqa:geometry_loss}
\end{aligned}
\end{equation}
\review{where $V$ denotes the non-rigid region extracted from stereo clean images by checking forward-backward consistency constraint \cite{zou2018df}:}
\begin{equation}
  \setlength\abovedisplayskip{4pt}
  \setlength\belowdisplayskip{4pt}
\begin{aligned}
  &|F^f(x) + F^b(x + F^f(x))|^2 < \\ \alpha_1 (|&F^f(x)|^2 + |F^b(x + F^f(x))|^2) + \alpha_2,
  \label{eq:consistency_constraint}
\end{aligned}
\end{equation}
\review{where $F$ is the rigid flow $F_c^{ri}$ here, and $x$ denotes the pixel index. $\alpha_1$ and $\alpha_2$ are two penalty parameters. The non-rigid mask $V$ is set to $1$ when the constraint Eq. \ref{eq:consistency_constraint} is violated, and $0$ otherwise.}

As for the relationship between depth and static weather (\emph{e.g.}, fog), atmospheric scattering model \cite{narasimhan2002vision} reveals that static weather affects the visual quality of the image along scene depth:
\begin{equation}
\begin{aligned}
  J = I \cdot t(D) + A \cdot (1-t(D)),
  \label{eqa:atmospheric_scattering_model}
\end{aligned}
\end{equation}
where $A$ denotes the predefined atmospheric light, $t(\cdot)$ is a decay function related depth $D$. We use Eq. \ref{eqa:atmospheric_scattering_model} to generate the synthetic foggy images corresponding to the left clean image, thus achieving an association between depth and static weather. To transfer pixel-wise motion knowledge, we enforce flow consistency loss:
\begin{equation}
\begin{aligned}
  \mathcal{L}^{consis}_{flowS} = ||F_{ss} - F_c||_1,
  \label{eq:flow_consistency_static}
\end{aligned}
\end{equation}
where $F_{ss}$ is the optical flow of degraded domain. Hence, depth physically associates the intrinsic motion and the degradation for the directional and explicit knowledge transfer.

\subsubsection{Warp Error Heterogeneous Adaptation}
The depth association homogeneous adaptation does benefit the optical flow estimation under static weather (\emph{e.g.}, fog), while it fails in dealing with optical flow under dynamic weather (\emph{e.g.}, rain). \review{The main reason is that, dynamic weather will bring in additional motion errors, resulting in that the homogeneous adaptation alone has the potential risk of misaligning the heterogeneous motion feature caused by dynamic degradation into the homogeneous intrinsic motion feature space. On the contrary, we aim to model the feature with discrepancy between clean and degraded domains, and explore a physical space to explicitly distinguish the motion caused by dynamic weather from the intrinsic motion of the scene.}

\noindent
\textbf{\review{Heterogeneous Boundary in Warp Error.}}
As shown in Fig. \ref{Degradation}, dynamic weather is an additional disturbance error added to the clean motion. Taking rain streak as an example, given a degraded image $Y$, we simply assume that the degradation is a linear additive decomposition model \cite{fu2017removing}:
\begin{equation}
  \setlength\abovedisplayskip{4pt}
  \setlength\belowdisplayskip{4pt}
\begin{aligned}
  Y = X + D,
  \label{eq:additive_model}
\end{aligned}
\end{equation}
where $X$ denotes the clean image, $D$ denotes the dynamic degradation component. We argue that the accuracy of the optical flow determines the degree of the alignment of the adjacent frames. Therefore, we use the optical flow to warp the two adjacent frames for the clean image and the degraded image as follows:
\begin{equation}
  \setlength\abovedisplayskip{4pt}
  \setlength\belowdisplayskip{4pt}
\begin{aligned}
  w(i) = I_1(i) - I_2(i+F(i)),
  \label{eq:warp_error}
\end{aligned}
\end{equation}
where $w$ is warp error operator, $I_1$, $I_2$ are adjacent frames, $F$ is optical flow, and $i$ is the pixel index. Then, according to the decomposition model Eq. \ref{eq:additive_model}, we can further model the warp error relationship between the clean images, degraded images and degradation component as follows:
\begin{equation}
  \setlength\abovedisplayskip{4pt}
  \setlength\belowdisplayskip{4pt}
\begin{aligned}
  w_y(i) - w_x(i) &=
  \resizebox{0.61\hsize}{!}{$({Y_1}(i) - {X_1}(i)) - ({Y_2}(i + F(i)) - {X_2}(i + F(i)))$} \\
   &= D_1(i) - D_2(i + F(i)) = w_d(i),
  \label{eq:heterogeneous_error}
\end{aligned}
\end{equation}
where $w_x$, $w_y$, $w_d$ denote the warp errors of the clean domain, degraded domain and dynamic degradation component. Eq. \ref{eq:heterogeneous_error} means that dynamic weather causes a significant boundary variance in the warp error space of optical flow between clean domain and degraded domain. \review{To further assess the warp error discrepancy between clean and degraded domains, we visualize the feature distributions of the clean image and the corresponding synthetic rainy image in the warp error space, as shown in Fig. \ref{WarpError}. We sample the patches of the intrinsic motion boundary in the clean image, the corresponding patches in the synthetic rainy image, and other patches related to rain streaks. We can observe that the motion boundary errors in the clean domain have a distribution similar to those in the degraded domain, whereas the distribution of the errors caused by dynamic weather is discrete and dissimilar to the motion boundary errors. This highlights the physical meaning of Eq. \ref{eq:heterogeneous_error}.}
Therefore, instead of directly modeling the errors of dynamic degradation component $w_d$, we can indirectly model the warp errors between clean and degraded domains, namely $w_y - w_x$. We use exponential function $\exp(\cdot)$ to transform Eq. \ref{eq:heterogeneous_error}:
\begin{equation}
  \setlength\abovedisplayskip{4pt}
  \setlength\belowdisplayskip{4pt}
\begin{aligned}
  \exp(w_d) = \exp(w_y - w_x) = \frac{\exp(w_y)}{\exp(w_x)}.
  \label{eq:exp_formula}
\end{aligned}
\end{equation}

Interestingly, we observe that Eq. \ref{eq:exp_formula} shares a similar formulation with contrastive learning \cite{chen2020simple}. Both of them are to measure the exclusive relationship between two domains. This motivates us to utilize a contrastive learning strategy to pull the clean errors of the intrinsic motion boundary together and push the degraded errors caused by dynamic weather away.

\begin{figure*}
  \centering
   \includegraphics[width=0.99\linewidth]{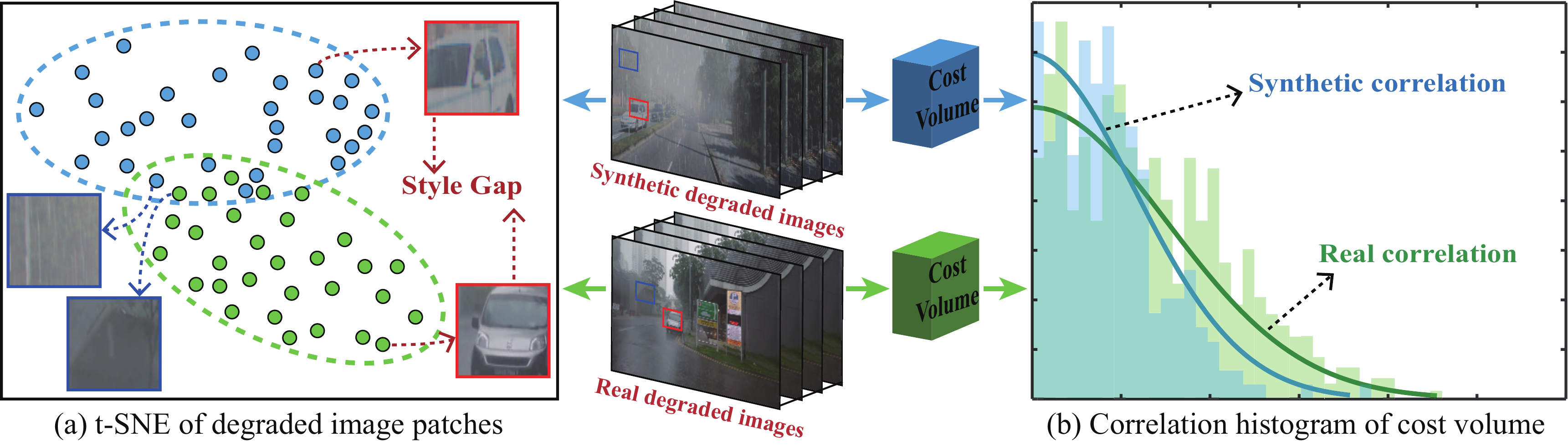}
   \caption{Visual distribution of synthetic and real degraded images. In (a) t-SNE of degraded image patches, the difference of degradation pattern is small, but there exists an obvious scene style gap between the two domains. In (b) correlation histogram of cost volume, the holistic correlation distributions are similar. This inspires us to align the homogeneous correlation distribution for the synthetic-real motion knowledge transfer.
   }
   \label{CostVolume}

\end{figure*}

\noindent
\textbf{Warp Error Contrastive Learning.}
To guarantee the robustness of the entire motion knowledge transfer from clean domain to degraded domain, we divide the knowledge transfer process for dynamic weather into two stages: motion consistency learning and warp error contrastive learning. Motion consistency learning is to directly transfer pixel-wise motion knowledge from clean domain to degraded domain for initializing the degraded flow model. Warp error contrastive learning is to further align the intrinsic motion boundaries and peel off the negative effect of degradation on optical flow in the warp error space. As for motion consistency learning stage, we use Eq. \ref{eq:additive_model} to synthesize the degraded images, and enforce the flow consistency loss on the optical flow result as follows:
\begin{equation}
\begin{aligned}
  \mathcal{L}^{consis}_{flowD} = ||F_{sd} - F_c||_1,
  \label{eq:flow_consistency}
\end{aligned}
\end{equation}
where $F_{sd}$ is the optical flow of synthetic degraded domain.
\review{Note that Eq. \ref{eq:flow_consistency} and Eq. \ref{eq:flow_consistency_static} jointly constitute the cross-domain flow consistency loss, namely $\mathcal{L}^{consis}_{flow} = \mathcal{L}^{consis}_{flowD} + \mathcal{L}^{consis}_{flowS}$, making the motion knowledge transfer process more robust to adverse weather.}
As for warp error contrastive learning stage, we further distinguish the image motion boundary from that of the erroneousness caused by the adverse degradation, such as rain streaks, thus refining the motion boundary. The key to contrastive learning is the sampling strategy of positives/negatives. For the positive, we propose an edge-aware sampling strategy by extracting the salient edges \cite{dollar2013structured} [green patches in Fig. \ref{Framework}] from the clean domain as the positive samples, in order to get rid of the most meaningless region without zero values. For the negative, we propose an entropy-aware sampling strategy by extracting the salient edges [pink patches in Fig. \ref{Framework}] from the degraded domain.
We sort the local entropy (degree of degradation) in descending order and choose the top $N$ patches. \review{Note that the negative patches would be sampled with different locations from the positives, so as to exclude the motion boundary of the clean image.} The advantage of the proposed sampling strategy over the conventional random sampling strategy is analyzed in the discussion. Therefore, we denote $f_{P_j^x}=E_s(w_{x_j})$, $f_{P_j^y}=E_s(w_{y_j})$ and $f_{N_i}=E_s(w_{y_i})$ for the output positive and negative features, respectively, where $E_s$ is the feature encoder of synthetic degraded domain. We enforce contrastive learning loss as follows:
\begin{equation}
\begin{aligned}
  \resizebox{0.89\hsize}{!}{$\mathcal{L}^{contra}_{warpErr} = \sum\nolimits_{j=1}^N \frac{exp(f_{P_{j}^x}\cdot f_{P_{j}^y} / \tau)}{exp(f_{P_{j}^x}\cdot f_{P_{j}^y} / \tau)+\sum\nolimits_{i=1}^N{exp(f_{N_{i}}\cdot f_{P_{j}^x}/\tau)}},$}
  \label{eqa:contra_loss}
\end{aligned}
\end{equation}
where $N$ is the sampling number of positives/negatives, and $\tau$ denotes the scale parameter. The warp error contrastive learning loss constrain the clean warp error patches $w_{x_j}$ at pixel location $j$ to be positive with the corresponding degraded warp error patches $w_{y_j}$ in comparison to other warp error patches $w_{y_i}$, so as to force the motion boundary of the degraded domain close to the motion boundary of clean domain in the same scene and get rid of degradation. In this way, motion consistency learning and warp error contrastive learning jointly promote the transfer of the scene intrinsic motion knowledge from clean to degraded domain, and peel off the degraded motion errors caused by dynamic disturbance.

\subsection{Synthetic-Real Motion Adaptation}
Depth association homogeneous motion adaptation for static weather and warp error heterogeneous boundary adaptation for dynamic weather can handle optical flow under adverse weather, thus bridging the clean-degraded knowledge transfer and providing a coarse optical flow model for synthetic degraded domain. However, these adaptations cannot help the synthetic-real knowledge transfer, and thus the proposed method may inevitably suffer artifacts when applied to real adverse weather scenes. To this end, we further explore a physical space for synthetic-real motion transfer.

\noindent
\textbf{\review{Homogeneous Correlation in Cost Volume.}}
\review{Unlike the pixel-wise motion knowledge transfer from the clean domain to the synthetic degraded domain, the synthetic degraded and real degraded domains are not from the same scene, thus making the pixel-wise knowledge transfer unsuitable. Instead, we propose to transform both synthetic and real degraded domains into a common feature space to address the distribution discrepancy between the two domains. The key challenge is identifying this feature space. According to our knowledge, optical flow is a task of modeling the mapping relationship from visual features to motion features. This drives us to visualize the feature distribution discrepancy between synthetic degraded and real degraded domains in both visual space and motion space, as shown in Fig. \ref{CostVolume}.} For the visual feature distribution, degradation pattern discrepancy between synthetic and real degraded images is small in Fig. \ref{CostVolume} (a), but there exists an obvious synthetic-real style gap that restricts the optical flow performance under real adverse weather scenes. For the motion feature distribution, we choose cost volume as the motion feature space, which can physically measure the similarity of two adjacent frames, not limited by scene difference. We transform the synthetic and real degraded images to cost volume space, and visualize the correlation distributions of synthetic and real degraded images via the histogram in Fig. \ref{CostVolume} (b). We observe that both the domains share a similar statistical histogram of cost volume. This motivates us to take correlation distribution as the homogeneous feature between synthetic and real domains, achieving the synthetic-real explicit motion knowledge transfer via feature distribution alignment.

\begin{figure*}
  \centering
   \includegraphics[width=0.99\linewidth]{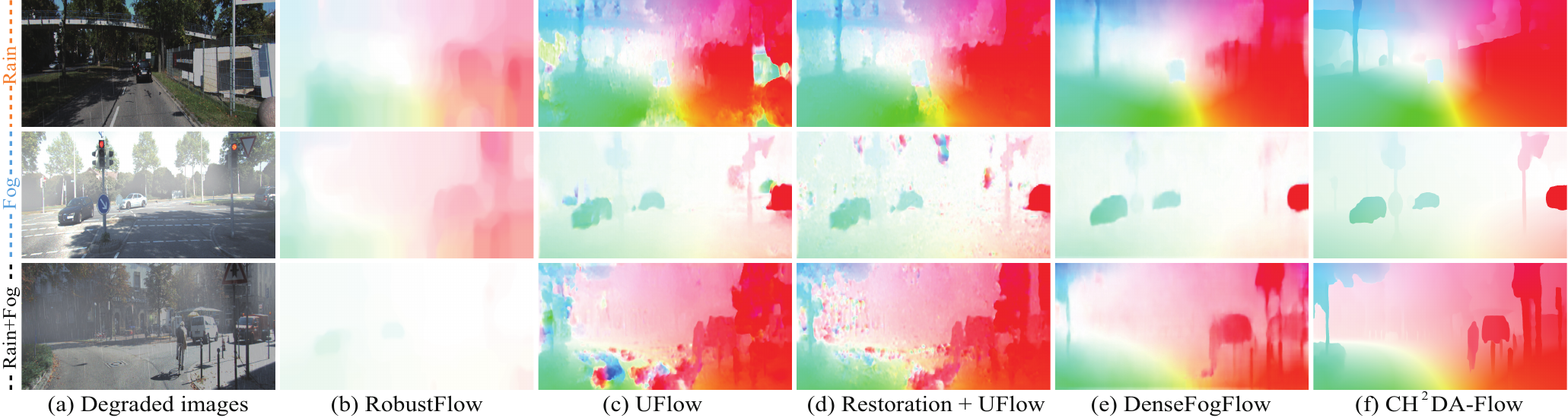}
   \caption{Visual comparison of optical flows on synthetic Weather-KITTI2015 dataset. (a) Degraded images (\emph{e.g.}, rain, fog and rain+fog). Optical flow estimated by (b) RobustFlow, (c) UFlow, (d) Restoration + UFlow, (e) DenseFogFlow, (f) CH$^2$DA-Flow. Compared with other competing methods, the proposed CH$^2$DA-Flow achieves the state-of-the-art performance with a clearer motion boundary.
   }
   \label{Synthetic}

\end{figure*}

\begin{table*}
  \setlength\tabcolsep{2pt}
\centering
\renewcommand\arraystretch{1.1}
\caption{Quantitative comparisons on synthetic Weather-KITTI2015 datasets, including Rain-KITTI2015 and Fog-KITTI2015. ``--'' denotes the original version of optical flow method without any preprocessing. ``RLNet / AECR-Net +'' denotes the image preprocessing. The best results are highlighted by bold.}
\begin{tabular}{cc|cccccccccc}

\Xhline{1pt}
\multicolumn{2}{c|}{Paradigm}&
\multicolumn{8}{c|}{Direct estimation}&
\multicolumn{2}{c}{Knowledge transfer}\\

\hline
\multicolumn{2}{c|}{\multirow{2}{*}{Method}}& \multicolumn{1}{c|}{\multirow{2}{*}{\makecell{Robust\\Flow}}} &
\multicolumn{3}{c|}{UFlow} & \multicolumn{3}{c|}{Selflow} &
\multicolumn{1}{c|}{\multirow{2}{*}{SMURF}} & \multicolumn{1}{c|}{\multirow{2}{*}{\makecell{DenseFog\\Flow}}} & \multirow{2}{*}{CH$^2$DA-Flow}\\

\cline{4-6}\cline{7-9}
\multicolumn{2}{c|}{}& \multicolumn{1}{c|}{} & \multicolumn{1}{c|}{--} & \multicolumn{1}{c|}{RLNet +}& \multicolumn{1}{c|}{AECR-Net +}&  \multicolumn{1}{c|}{--}& \multicolumn{1}{c|}{RLNet-Net +}& \multicolumn{1}{c|}{AECR-Net +}  & \multicolumn{1}{c|}{}& \multicolumn{1}{c|}{} &\\
 \hline
\multicolumn{1}{c|}{\multirow{2}{*}{Rain-KITTI}} & EPE & \multicolumn{1}{c|}{23.40}& \multicolumn{1}{c|}{8.54} &\multicolumn{1}{c|}{7.63}& \multicolumn{1}{c|}{--}& \multicolumn{1}{c|}{10.78}& \multicolumn{1}{c|}{7.04}& \multicolumn{1}{c|}{--}& \multicolumn{1}{c|}{7.45}& \multicolumn{1}{c|}{6.96} & \textbf{5.35} \\
\cline{2-12}
\multicolumn{1}{c|}{}& F1-all & \multicolumn{1}{c|}{81.12\%} & \multicolumn{1}{c|}{47.85\%} & \multicolumn{1}{c|}{42.07\%}& \multicolumn{1}{c|}{--} & \multicolumn{1}{c|}{52.18\%} & \multicolumn{1}{c|}{41.35\%} & \multicolumn{1}{c|}{--} & \multicolumn{1}{c|}{43.74\%}  & \multicolumn{1}{c|}{41.02\%} & \textbf{30.29\%} \\

\hline
\multicolumn{1}{c|}{\multirow{2}{*}{Fog-KITTI}} & EPE & \multicolumn{1}{c|}{25.32}& \multicolumn{1}{c|}{16.55} &\multicolumn{1}{c|}{--}& \multicolumn{1}{c|}{12.16}& \multicolumn{1}{c|}{15.84}& \multicolumn{1}{c|}{--}& \multicolumn{1}{c|}{11.21}& \multicolumn{1}{c|}{11.56}& \multicolumn{1}{c|}{8.03} & \textbf{5.68} \\
\cline{2-12}
\multicolumn{1}{c|}{}& F1-all & \multicolumn{1}{c|}{85.77\%} & \multicolumn{1}{c|}{62.84\%} & \multicolumn{1}{c|}{--}& \multicolumn{1}{c|}{53.17\%} & \multicolumn{1}{c|}{58.81\%} & \multicolumn{1}{c|}{--} & \multicolumn{1}{c|}{50.25\%} & \multicolumn{1}{c|}{51.39\%}  & \multicolumn{1}{c|}{41.73\%} & \textbf{32.47\%} \\
\Xhline{1pt}
\end{tabular}
 \label{Quantitative_Synthetic_Rain}
\end{table*}

\noindent
\textbf{Correlation Distribution Holistic Alignment.}
Synthetic-real knowledge transfer process can be divided into two steps: feature holistic alignment and optical flow pseudo-supervised. In feature holistic alignment stage, we begin with two encoders $E_s$, $E_r$ for the synthetic degraded images $[J_t^s, J_{t+1}^s]$ and real degraded images $[J_t^r, J_{t+1}^r]$, respectively. We encode them to obtain the cost volumes $cv_s$, $cv_r$  with the warp operator. Furthermore, we randomly sample $M$ correlation values in the cost volumes $cv_s$, $cv_r$ to represent the holistic correlation distribution of cost volume normalized into $[0, 1]$ for both the domains. According to the range of correlation, we choose threshold values $[\delta_1, \delta_2, ..., \delta_{k-1}]$ to label the sampled correlation into $k$ classes, \emph{e.g.}, high correlation and low correlation. The correlation distribution $p$ is computed by:
\begin{equation}
\begin{aligned}
  p = \frac{m + 1}{M + k},
  \label{eq:distribution}
\end{aligned}
\end{equation}
where $m$ denotes the number of the sampled correlations of one category. \review{Note that we add an offset 1 to each category sampled correlation of Eq. \ref{eq:distribution} to ensure at least a single instance could be present in the real degraded domain.} Thus, to align the motion features between synthetic and real degraded domains, we minimize the correlation distribution distance between the two domains by enforcing Kullback-Leibler divergence:
\begin{equation}
  \setlength\abovedisplayskip{4pt}
  \setlength\belowdisplayskip{4pt}
\begin{aligned}
  \mathcal{L}^{kl}_{corr} = \sum\nolimits_{i=1}^k{p_{r,i}log\frac{p_{r,i}}{p_{s,i}}},
  \label{eq:kl_div}
\end{aligned}
\end{equation}
where $p_{s,i}$, $p_{r,i}$ denote the $i$ category sampled correlation distributions of the synthetic degraded domain and the real degraded domain, respectively. The aligned correlation distributions represent that both the domains could have similar optical flow estimation capabilities. Then, we decode the aligned correlation of cost volume to predict optical flow for real degraded images. In optical flow pseudo-supervised stage, to improve the robustness of knowledge transfer, we propose a self-supervised training strategy that attempts to transfer motion knowledge from synthetic to real domain at the optical flow field level. We feed the real degraded images $[J_t^r, J_{t+1}^r]$ to the flow network of the synthetic degraded domain, which outputs the optical flow as the pseudo-labels $F_{pse}$ for the optical flow model of real degraded domain (seeing the red arrow in Fig. \ref{Framework}). We then impose a self-supervised loss on the optical flow $F_r$ estimated by the flow network of the real degraded domain:
\begin{equation}
  \setlength\abovedisplayskip{5pt}
  \setlength\belowdisplayskip{4pt}
\begin{aligned}
  \mathcal{L}^{self}_{flow} = \sum\nolimits||F_r - F_{pse}||_1.
  \label{eq:self_flow}
\end{aligned}
\end{equation}

During the training process, the encoder $E_r(f;\theta_r)$ of the real degraded domain is updated with the encoder $E_s(f;\theta_s)$ of the synthetic degraded domain using the exponential moving average (EMA) mechanism as follows:
\begin{equation}
  \setlength\abovedisplayskip{4pt}
  \setlength\belowdisplayskip{4pt}
\begin{aligned}
  \theta_r \leftarrow \theta_r \cdot \lambda + \theta_s \cdot (1 - \lambda),
  \label{eq:EMA}
\end{aligned}
\end{equation}
where ${\lambda}$ controls the window of EMA and is often close to 1.0.
\review{The proposed homogeneous correlation adaptation can effectively distill motion knowledge of the synthetic degraded domain to the real degraded domain within the similar cost volume correlation feature and intrinsic optical flow dimensions.}

\begin{figure*}
  \centering
   \includegraphics[width=0.99\linewidth]{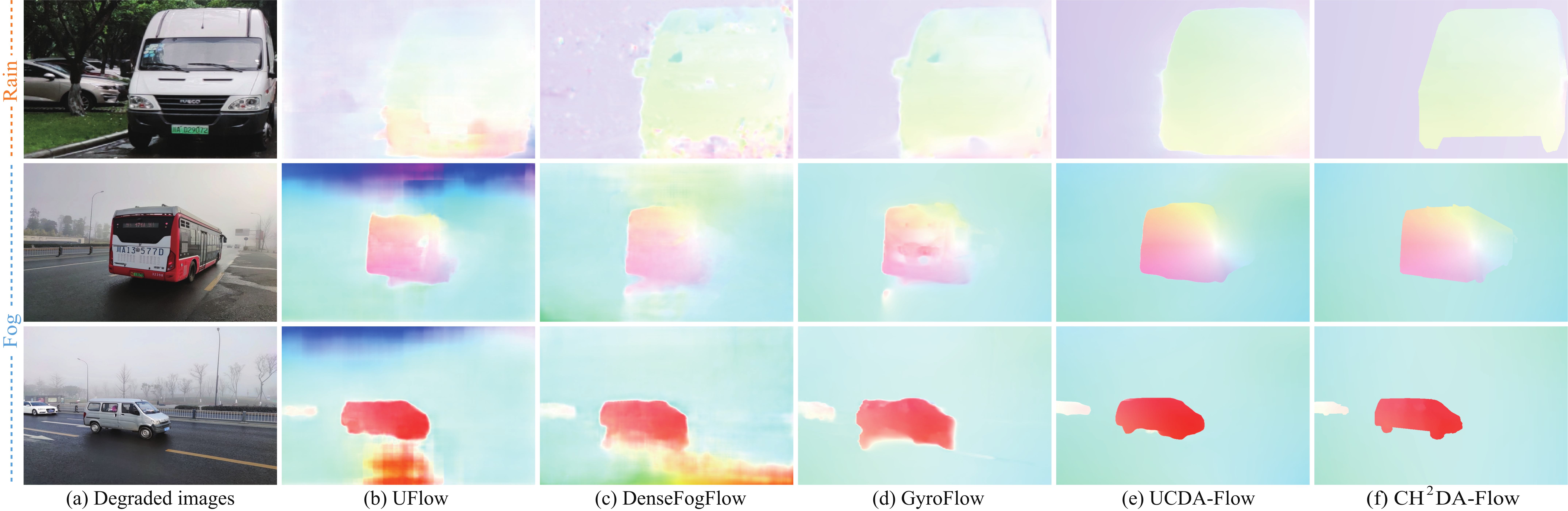}
   \caption{Visual comparison of optical flows on real Weather-GOF dataset. (a) Degraded images (\emph{e.g.}, rain, fog). Optical flow estimated by (b) UFlow, (c) DenseFogFlow, (d) GyroFlow, (e) UCDA-Flow, (f) CH$^2$DA-Flow. Compared with other competing methods, the previous version UCDA-Flow and the proposed CH$^2$DA-Flow both perform better with sharp boundaries and smooth background, while the motion structure of CH$^2$DA-Flow is clearer.
   }
   \label{Real}

\end{figure*}

\begin{table*}
  \setlength\tabcolsep{1pt}
\centering
\renewcommand\arraystretch{1.1}
\caption{Quantitative comparisons on real public optical flow datasets under adverse weather, including Weather-GOF and DenseFog. ``--'' denotes the original version of optical flow method without any preprocessing. ``ssl'' denotes being trained on target real datasets via self-supervised learning \cite{stone2021smurf}.}
\begin{tabular}{cc|ccccccccccccc}

\Xhline{1pt}
\multicolumn{2}{c|}{Paradigm}&
\multicolumn{7}{c|}{Direct estimation}&
\multicolumn{5}{c}{Knowledge transfer}\\
\hline
\multicolumn{2}{c|}{\multirow{2}{*}{Method}}& \multicolumn{1}{c|}{\multirow{2}{*}{\makecell{Robust\\Flow}}} & \multicolumn{1}{c|}{\multirow{2}{*}{UFlow}} &
\multicolumn{1}{c|}{\multirow{2}{*}{SMURF}} &
\multicolumn{2}{c|}{PWC-Net} & \multicolumn{2}{c|}{GMA} & \multicolumn{1}{c|}{\multirow{2}{*}{GyroFlow}} &
\multicolumn{1}{c|}{\multirow{2}{*}{\makecell{DenseFog\\Flow}}} &
\multicolumn{1}{c|}{\multirow{2}{*}{AttackFlow}} &
\multicolumn{1}{c|}{\multirow{2}{*}{\makecell{HMBA\\-Flow}}} &
\multicolumn{1}{c|}{\multirow{2}{*}{\makecell{UCDA\\-Flow}}} &
\multicolumn{1}{c}{\multirow{2}{*}{\makecell{CH$^2$DA\\-Flow}}} \\

\cline{6-9}

\multicolumn{2}{c|}{}& \multicolumn{1}{c|}{} & \multicolumn{1}{c|}{}&
\multicolumn{1}{c|}{}&\multicolumn{1}{c|}{--}&\multicolumn{1}{c|}{ssl} &\multicolumn{1}{c|}{--} & \multicolumn{1}{c|}{ssl}& \multicolumn{1}{c|}{}&  \multicolumn{1}{c|}{}&
\multicolumn{1}{c|}{} & \multicolumn{1}{c|}{} & \multicolumn{1}{c|}{}& \multicolumn{1}{c}{}\\

\hline
\multicolumn{1}{c|}{\multirow{2}{*}{Rain-GOF}} & EPE & \multicolumn{1}{c|}{7.10}& \multicolumn{1}{c|}{2.01} &\multicolumn{1}{c|}{1.72}& \multicolumn{1}{c|}{2.85}& \multicolumn{1}{c|}{3.01}& \multicolumn{1}{c|}{1.45}& \multicolumn{1}{c|}{1.39}& \multicolumn{1}{c|}{1.07}& \multicolumn{1}{c|}{1.32} &\multicolumn{1}{c|}{1.25}&
\multicolumn{1}{c|}{0.91}&\multicolumn{1}{c|}{0.90}& \textbf{0.74} \\

\cline{2-15}
\multicolumn{1}{c|}{}& F1-all & \multicolumn{1}{c|}{53.35\%} & \multicolumn{1}{c|}{20.25\%} & \multicolumn{1}{c|}{16.11\%}& \multicolumn{1}{c|}{21.39\%} & \multicolumn{1}{c|}{22.55\%} & \multicolumn{1}{c|}{12.21\%} & \multicolumn{1}{c|}{12.17\%} & \multicolumn{1}{c|}{9.94\%}  & \multicolumn{1}{c|}{11.59\%} & \multicolumn{1}{c|}{10.84\%}& \multicolumn{1}{c|}{8.07\%}& \multicolumn{1}{c|}{8.05\%}& \textbf{7.05\%} \\

\hline
\multicolumn{1}{c|}{\multirow{2}{*}{Fog-GOF}} & EPE & \multicolumn{1}{c|}{12.25}& \multicolumn{1}{c|}{2.97} &\multicolumn{1}{c|}{1.92}& \multicolumn{1}{c|}{5.99}& \multicolumn{1}{c|}{6.02}& \multicolumn{1}{c|}{1.63}& \multicolumn{1}{c|}{1.69}& \multicolumn{1}{c|}{0.95}& \multicolumn{1}{c|}{1.78} &\multicolumn{1}{c|}{1.40}&
\multicolumn{1}{c|}{0.84}&\multicolumn{1}{c|}{0.81}& \textbf{0.72} \\

\cline{2-15}
\multicolumn{1}{c|}{}& F1-all & \multicolumn{1}{c|}{80.93\%} & \multicolumn{1}{c|}{30.82\%} & \multicolumn{1}{c|}{17.86\%}& \multicolumn{1}{c|}{41.02\%} & \multicolumn{1}{c|}{45.25\%} & \multicolumn{1}{c|}{14.25\%} & \multicolumn{1}{c|}{15.11\%} & \multicolumn{1}{c|}{9.13\%}  & \multicolumn{1}{c|}{16.41\%} & \multicolumn{1}{c|}{11.95\%}& \multicolumn{1}{c|}{7.46\%}& \multicolumn{1}{c|}{7.18\%}& \textbf{6.85\%} \\

\hline
\multicolumn{1}{c|}{\multirow{2}{*}{DenseFog}} & EPE & \multicolumn{1}{c|}{13.48}& \multicolumn{1}{c|}{6.21} &\multicolumn{1}{c|}{5.84}& \multicolumn{1}{c|}{6.15}& \multicolumn{1}{c|}{6.12}& \multicolumn{1}{c|}{3.68}& \multicolumn{1}{c|}{3.81}& \multicolumn{1}{c|}{--}& \multicolumn{1}{c|}{4.32} &\multicolumn{1}{c|}{3.50}&
\multicolumn{1}{c|}{3.02}&\multicolumn{1}{c|}{2.94}& \textbf{2.82} \\

\cline{2-15}
\multicolumn{1}{c|}{}& F1-all & \multicolumn{1}{c|}{79.31\%} & \multicolumn{1}{c|}{62.45\%} & \multicolumn{1}{c|}{55.78\%}& \multicolumn{1}{c|}{60.70\%} & \multicolumn{1}{c|}{56.31\%} & \multicolumn{1}{c|}{33.18\%} & \multicolumn{1}{c|}{35.20\%} & \multicolumn{1}{c|}{--}  & \multicolumn{1}{c|}{41.26\%} & \multicolumn{1}{c|}{31.54\%}& \multicolumn{1}{c|}{28.94\%}& \multicolumn{1}{c|}{28.67\%}& \textbf{27.90\%} \\

\Xhline{1pt}
\end{tabular}
 \label{Quantitative_Real}
\end{table*}

\subsection{Optimization and Implementation Details}
Consequently, the total objective for the proposed framework is written as follows,
\begin{equation}
  \setlength\abovedisplayskip{4pt}
  \setlength\belowdisplayskip{4pt}
\begin{aligned}
  \mathcal{L} &= \lambda_1\mathcal{L}^{pho}_{flow} + \lambda_2\mathcal{L}_{depth} + \lambda_3\mathcal{L}^{geo}_{flow} +\lambda_4\mathcal{L}^{consis}_{flow} \\
  & + \lambda_5\mathcal{L}^{contra}_{warpErr} + \lambda_6\mathcal{L}^{self}_{flow} + \lambda_7\mathcal{L}^{kl}_{corr},
  \label{eq:total_loss}
\end{aligned}
\end{equation}
where the first term is to initialize the optical flow model of clean domain, making the whole framework to have the ability of learning the intrinsic motion patterns. The second, third and fourth terms aim to directionally transfer global motion knowledge along depth from clean domain to synthetic static degraded domain. The intention of the fourth and fifth terms is to use warp error to explicitly transfer local motion boundary knowledge from clean domain to synthetic dynamic degraded domain, and inhibit the negative impact of degradation on optical flow. The last two terms focus on transferring motion distribution knowledge in the cost volume feature space from synthetic degraded domain to real degraded domain. $[\lambda_1, \lambda_2, \lambda_3, \lambda_4, \lambda_5, \lambda_6, \lambda_7]$ are the weight parameters that control the importance of related losses, which are empirically set as $[1.0, 1.0, 0.1, 1.0, 0.1, 1.0, 0.1]$. Besides, we set the sampling number $N$ of positive/negative in contrastive learning as 100, the sampling number $M$ of correlation as 1000, and the number $k$ of categories as 10. The classification threshold values $\delta$ are set linearly from $[0, 1]$. The weight $\lambda$ of the EMA is 0.99.

\review{It should be emphasized that we only use the proposed framework to learn the optical flow model in the training phase, but directly infer the optical flow model in the inference phase.} During the training phase, we only need three steps. First, we train the optical flow network and disp network of clean domain via $\mathcal{L}^{pho}_{flow}$, $\mathcal{L}_{depth}$ and $\mathcal{L}^{geo}_{flow}$ for initializing the ability of modeling the motion of the whole framework. Second, we then train the optical flow network of synthetic degraded domain via $\mathcal{L}^{consis}_{flow}$ and $\mathcal{L}^{contra}_{warpErr}$ for closing the motion discrepancy between clean and degraded domains.
Finally, we further update the weights of the optical flow model of real degraded domain via $\mathcal{L}^{self}_{flow}$ and $\mathcal{L}^{kl}_{corr}$ for aligning the motion distributions between synthetic and real domains with 1000 epochs and 0.0005 learning rate. Under the unified alternate optimization framework, the proposed cumulative knowledge transfer method can progressively transfer motion knowledge from clean domain to synthetic degraded domain, and then to real degraded domain. During the inference phase, the testing model only needs the flow model of real degraded domain, including the encoder $E_r$, warp, cost volume and flow decoder, achieving an end-to-end estimation of adverse weather optical flow.

\section{Experiments}
\subsection{Datasets and Experimental Settings}
\textbf{Dataset.}
We take the KITTI2015 \cite{menze2015object} dataset as the representative clean scene. We conduct extensive comparison and ablation experiments on one synthetic and three real degraded datasets.

\noindent
$\bullet$ \textbf{Weather-KITTI2015.}
We construct a synthetic adverse weather KITTI dataset with rain (named as Rain-KITTI2015) on images of KITTI2015 via linearly additive model Eq. \ref{eq:additive_model} and fog (named as Fog-KITTI2015) on image of KITTI2015 via atmospheric scattering model Eq. \ref{eqa:atmospheric_scattering_model}. We select 8400 images of Fog/Rain-KITTI2015 datasets for training and 400 images for testing.

\noindent
$\bullet$ \textbf{Weather-GOF.}
GOF \cite{li2021gyroflow} is a public real degraded dataset containing four different scenes with synchronized gyro readings, such as regular scenes, rainy scenes, foggy scenes and nighttime scenes. We choose rainy images and foggy images of GOF to compose a new adverse weather dataset, namely Weather-GOF, of which 1000 images for training and 105 images for testing.

\noindent
$\bullet$ \textbf{DenseFog.}
We seek real foggy dataset with manually handcrafted motion labels collected from DenseFogFlow \cite{yan2020optical}, namely DenseFog, of which 2346 images for training and 200 for testing.

\begin{figure*}
  \centering
   \includegraphics[width=0.99\linewidth]{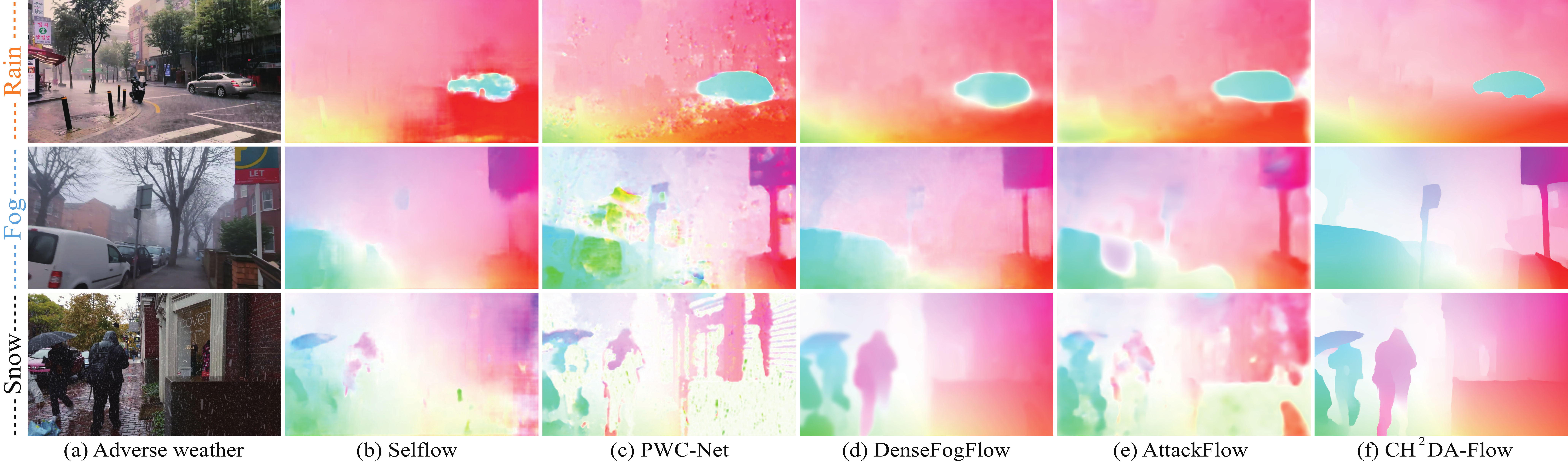}
   \caption{Visual comparison of optical flows on the proposed Real-Weather World dataset. (a) Degraded images (\emph{e.g.}, rain, fog and snow). Optical flow estimated by (b) Selflow, (c) PWC-Net, (d) DenseFogFlow, (e) AttackFlow, (f) CH$^2$DA-Flow. Compared with other competing methods, the proposed CH$^2$DA-Flow generalizes well for various adverse weather conditions with fewer artifacts and sharp boundaries.
   }
   \label{Unseen}

\end{figure*}

\noindent
$\bullet$ \textbf{Real-Weather World.}
In the existing real adverse weather optical flow datasets, the scenes are relatively single, and independent moving objects are few. \review{In this work, we propose an optical flow dataset that covers multiple independent moving object samples under various adverse weather conditions (\emph{e.g.}, rain, fog and snow)}. \review{For the collection, we obtain the video sequences from \emph{Youtube} and camera, and then manually split them into paired adjacent images according to weather types and scene diversity. For the ground truth, with the help of the human-assisted motion annotation method \cite{liu2008human}, we first segment the mask of the moving object, and then manually label key points in adjacent images for the moving object, thus generating the corresponding motion displacement as the ground truth. Note that the proposed dataset is not used for the model training, but for the generalization comparison of different competing methods during the inference phase.}

\begin{table}
  \setlength\tabcolsep{1.5pt}
\centering
\renewcommand\arraystretch{1.1}
\caption{\review{Quantitative comparisons on the proposed real optical flow dataset under various adverse weather conditions, including rain, fog, and snow.}}
\begin{tabular}{cc|cccccc}
\Xhline{1pt}
\multicolumn{2}{c|}{Paradigm}&
\multicolumn{3}{c|}{Direct estimation}&
\multicolumn{3}{c}{Knowledge transfer}\\
\hline
\multicolumn{2}{c|}{Method}&  \multicolumn{1}{c|}{Selflow} &
\multicolumn{1}{c|}{\makecell{PWC\\-Net}} & \multicolumn{1}{c|}{GMA} & \multicolumn{1}{c|}{\makecell{Dense\\FogFlow}} & \multicolumn{1}{c|}{\makecell{Attack\\Flow}} & \multicolumn{1}{c}{Ours} \\

\hline
\multicolumn{1}{c|}{\multirow{2}{*}{\makecell{Rain}}}& \multicolumn{1}{c|}{EPE} &  \multicolumn{1}{c|}{6.76} & \multicolumn{1}{c|}{6.95}& \multicolumn{1}{c|}{3.75}&  \multicolumn{1}{c|}{3.91}& \multicolumn{1}{c|}{3.14} & \multicolumn{1}{c}{\textbf{2.34}}\\
\cline{2-8}
\multicolumn{1}{c|}{}& \multicolumn{1}{c|}{F1-all} & \multicolumn{1}{c|}{53.30\%} & \multicolumn{1}{c|}{56.52\%}& \multicolumn{1}{c|}{37.47\%}&  \multicolumn{1}{c|}{36.64\%}& \multicolumn{1}{c|}{29.65\%} & \multicolumn{1}{c}{\textbf{22.61\%}}\\
\hline
\multicolumn{1}{c|}{\multirow{2}{*}{\makecell{Fog}}}& \multicolumn{1}{c|}{EPE}  & \multicolumn{1}{c|}{7.04} & \multicolumn{1}{c|}{6.80}& \multicolumn{1}{c|}{3.64}&  \multicolumn{1}{c|}{3.98}& \multicolumn{1}{c|}{3.29} & \multicolumn{1}{c}{\textbf{2.45}}\\
\cline{2-8}
\multicolumn{1}{c|}{}& \multicolumn{1}{c|}{F1-all} & \multicolumn{1}{c|}{61.24\%} & \multicolumn{1}{c|}{52.48\%}& \multicolumn{1}{c|}{34.08\%}&  \multicolumn{1}{c|}{36.75\%}& \multicolumn{1}{c|}{31.46\%} & \multicolumn{1}{c}{\textbf{23.80\%}}\\

\hline
\multicolumn{1}{c|}{\multirow{2}{*}{\makecell{Snow}}}& \multicolumn{1}{c|}{EPE} & \multicolumn{1}{c|}{6.65} & \multicolumn{1}{c|}{6.17}& \multicolumn{1}{c|}{3.60}&  \multicolumn{1}{c|}{3.73}& \multicolumn{1}{c|}{2.94} & \multicolumn{1}{c}{\textbf{2.37}}\\
\cline{2-8}
\multicolumn{1}{c|}{}& \multicolumn{1}{c|}{F1-all} & \multicolumn{1}{c|}{54.22\%} & \multicolumn{1}{c|}{51.35\%}& \multicolumn{1}{c|}{32.83\%}&  \multicolumn{1}{c|}{32.91\%}& \multicolumn{1}{c|}{28.55\%} & \multicolumn{1}{c}{\textbf{22.56\%}}\\
\Xhline{1pt}
\end{tabular}
 \label{Quantitative_Unseen}
\end{table}

\noindent
\textbf{Comparison Methods.}
We choose five competing methods optimization-based RobustFlow \cite{li2018robust}, semi-supervised DenseFogFlow \cite{yan2020optical}, weakly-supervised GyroFlow \cite{li2021gyroflow}, supervised AttackFlow \cite{Schmalfuss_2023_ICCV} and unsupervised HMBA-Flow \cite{Zhou_aaai} which are designed for adverse weather optical flow. Moreover, we also select several state-of-the-art supervised (GMA \cite{jiang2021transformer} and PWC-Net \cite{sun2018pwc}) and unsupervised (SMURF \cite{stone2021smurf}, UFlow \cite{jonschkowski2020matters} and Selflow \cite{liu2019selflow}) optical flow methods designed for clean scenes. Furthermore, we also compare the proposed method with the previous version UCDA-Flow \cite{zhou2023unsupervised}.
For a fair comparison, we train all the competing methods on the same dataset settings. The weakly-supervised, semi-supervised, and unsupervised methods are first trained on KITTI2015 for initialization and re-trained on the target degraded dataset. The supervised methods are first trained on KITTI2015 in a supervised manner, and then trained on target degraded dataset via self-supervised learning \cite{stone2021smurf}, denoted with ``ssl'' in Table \ref{Quantitative_Real}.
As for the comparison on Weather-KITTI2015, we design two different strategies for competing methods. The first is that we directly train the competing methods on degraded images. The second is to perform the image restoration first via derain (\emph{e.g.}, RLNet \cite{9577602}) and defog (\emph{e.g.}, AECR-Net \cite{wu2021contrastive}) approaches, and then we train the competing methods on the deraining (named as RLNet+) and defogfing (named as AECR-Net+) results. \review{Note that this comparison is to demonstrate the limited effect of image restoration on adverse weather optical flow.
In addition, we also quantitatively and qualitatively compare the generalization of all the competing methods for various adverse weather conditions on the proposed Real-Weather World dataset.}

\noindent
\textbf{Evaluation Metrics.}
We choose the average end-point error (EPE \cite{dosovitskiy2015flownet}) and the lowest percentage of erroneous pixels (Fl-all \cite{menze2015object}) as the evaluation metrics for the quantitative evaluation. The smaller the index is, the better the predicted result is.

\subsection{Comparisons with State-of-the-arts}
\textbf{Comparisons on Synthetic Images.}
In Table \ref{Quantitative_Synthetic_Rain}, we conduct comparison experiments on the synthetic Rain-KITTI2015 and Fog-KITTI2015 datasets, respectively. Since the proposed framework is trained in an unsupervised manner, we choose the unsupervised methods for a fair comparison. We have three key observations. First, existing optical flow methods cannot directly estimate accurate optical flow from degraded images. This is because adverse weather degradation breaks the brightness constancy and gradient continuity assumptions which optical flow heavily relies on. Second, the preprocess strategy of deraining (\emph{e.g.}, RLNet + UFlow) and dofogging (\emph{e.g.}, AECR-Net + UFlow) approaches is positive to optical flow estimation. However, since these image restoration methods are not designed for optical flow, they can indeed enhance the spatial visual quality of images, while the residual degradation or excessive smoothing can also affect the temporal motion matching of adjacent frames, thus limiting the performance of optical flow estimation. Third, knowledge transfer methods, such as DenseFogFlow and the proposed method CH$^2$DA-Flow, can significantly improve optical flow on synthetic degraded images. The reason behind this is that knowledge transfer bypasses the difficulty of directly estimating optical flow on degraded images by transferring motion knowledge from clean domain to degraded domain, largely preserving the ability to learn intrinsic motion patterns of the scenes. It is worth mentioning that in Fig. \ref{Synthetic}, the visual results of DenseFogFlow are inferior to the proposed method. The key difference is that the transfer process of the DenseFogFlow is relatively implicit, while the proposed method CH$^2$DA-Flow explicitly explores the physical features between clean and degraded domains to guide knowledge transfer.

\noindent
\textbf{Comparisons on Real Datasets.}
In Table \ref{Quantitative_Real} and Fig. \ref{Real}, we quantitatively and qualitatively compare the optical flow performance of different methods under real adverse weather (\emph{e.g.}, dynamic rain and static fog), where all the comparison methods are trained using unsupervised or self-supervised strategies for fairness. Optimization-based RobustFlow almost failed to work due to handcraft features being too ideal to represent real scenes. Unsupervised and supervised methods include many artifacts, since degradation breaks the optical flow basic assumption, exacerbating motion feature mismatching. In contrast, direct knowledge transfer methods significantly improve optical flow performance. However, these methods do utilize paired clean and synthetic images to bridge the clean-degraded transfer, while neglecting the synthetic-real style gap, and thus are erroneous when applied for real degraded images. GyroFlow resorts to gyroscope to model the motion of background for real degraded images, while failing with independent moving objects. Therefore, the proposed cumulative knowledge transfer frameworks (\emph{e.g.}, UCDA-Flow and CH$^2$DA-Flow) introduce the synthetic degraded domain as an intermediate bridge to close the clean-degraded gap and synthetic-real gap, thereby achieving the state-of-the-art performance. In addition, the previous version UCDA-Flow exhibits limited capability in estimating optical flow under dynamic weather, while the proposed method could comprehensively address the optical flow estimation under different adverse weather conditions.

\begin{figure}
  \centering
   \includegraphics[width=0.99\linewidth]{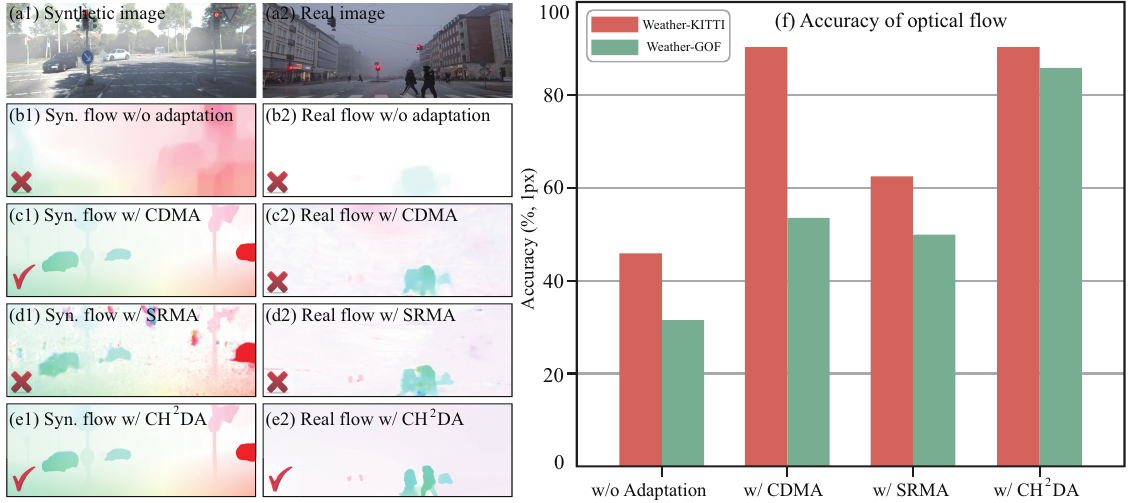}
   \caption{Effectiveness of cumulative domain adaptation framework. (a) Synthetic and real degraded images. Optical flow estimated (b) without any adaptation, (c) with clean-degraded motion adaptation (CDMA), (d) with synthetic-real motion adaptation (SRMA), (e) with CH$^2$DA. (f) Accuracy of optical flow. CDMA and SRMA jointly ensure the progressive transfer of motion knowledge from clean domain to real degraded domain.
   }
   \label{Ablation_Cumulative}

\end{figure}

\begin{table}
    \setlength\tabcolsep{1pt}
  \centering
  \caption{Choice of adaptation technique during knowledge transfer. Implicit feature denotes the feature output from flow encoder. Homogeneous and heterogeneous adaptations play an important role in knowledge transfer.}
  \renewcommand\arraystretch{1.1}
  \begin{tabular}{cc|cccc}
    \Xhline{1pt}
    \multicolumn{1}{c|}{\multirow{2}{*}{\makecell{Adaptation\\framework}}}&
    \multicolumn{1}{c|}{\multirow{2}{*}{Adaptation technique}}&
    \multicolumn{2}{c|}{Weather-GOF} &
    \multicolumn{2}{c}{DenseFog} \\
    \cline{3-6}
    \multicolumn{1}{c|}{}&\multicolumn{1}{c|}{} & \multicolumn{1}{c|}{EPE} & \multicolumn{1}{c|}{Fl-all} & \multicolumn{1}{c|}{EPE} & \multicolumn{1}{c}{Fl-all}\\
    \hline
    \multicolumn{2}{c|}{w/o knowledge transfer} & \multicolumn{1}{c|}{2.94} & \multicolumn{1}{c|}{30.85\%} & \multicolumn{1}{c|}{5.74} & \multicolumn{1}{c}{55.07\%}\\
    \hline
    \multicolumn{1}{c|}{\multirow{4}{*}{\makecell{Cumulative\\transfer}}} &
    \multicolumn{1}{c|}{w/ implicit feature} & \multicolumn{1}{c|}{1.21} & \multicolumn{1}{c|}{10.44\%} & \multicolumn{1}{c|}{3.25} & \multicolumn{1}{c}{30.18\%}\\
    \cline{2-6}
    \multicolumn{1}{c|}{} & \multicolumn{1}{c|}{w/ homogeneous} & \multicolumn{1}{c|}{0.82} & \multicolumn{1}{c|}{7.36\%} & \multicolumn{1}{c|}{2.94} & \multicolumn{1}{c}{28.67\%}\\
    \cline{2-6}
    \multicolumn{1}{c|}{} & \multicolumn{1}{c|}{w/ heterogeneous} & \multicolumn{1}{c|}{1.03} & \multicolumn{1}{c|}{9.45\%} & \multicolumn{1}{c|}{3.11} & \multicolumn{1}{c}{29.66\%}\\
    \cline{2-6}
    \multicolumn{1}{c|}{} & \multicolumn{1}{c|}{w/ homo-heterogeneous} & \multicolumn{1}{c|}{\textbf{0.72}} & \multicolumn{1}{c|}{\textbf{6.90\%}} & \multicolumn{1}{c|}{\textbf{2.82}} & \multicolumn{1}{c}{\textbf{27.90\%}} \\
    \Xhline{1pt}
  \end{tabular}
   \label{tab:Ablation_HomeHete}
\end{table}

\noindent
\review{\textbf{Generalization for Adverse Weather.}}
\review{To compare the generalization of different methods for various adverse weather conditions, we test the competing methods on the proposed dataset in Table \ref{Quantitative_Unseen} and Fig. \ref{Unseen}.} We have two conclusions. First, the generalization capability of knowledge transfer methods is significantly superior to those directly estimating methods in degraded scenes. The main reason is that directly estimating methods are prone to learning the degraded motion patterns from degraded images, such as dynamic weather, while knowledge transfer methods learn intrinsic motion patterns from clean domain and then transfer them to degraded domain, thus avoiding degradation interference. Second, the generalization capability of direct knowledge transfer methods like DenseFogFlow and AttackFlow is lower than that of the proposed framework under different weather conditions. Knowledge transfer from clean to real degraded domain faces a difficulty: a large domain gap. \review{Direct transfer methods neglect the large inter-domain gap and their transfer process is implicit, thus limiting the transfer ability. On the contrary, the proposed cumulative framework introduces an intermediate bridge to close the large domain gap between clean and real degraded domains.} Moreover, we further design the homogeneous and heterogeneous adaptation techniques to directionally guide the knowledge transfer of intrinsic scene motion, thus making the transfer process more robust to degradation and improving the generalization of the proposed method for real adverse weather.

\begin{table}
    \setlength\tabcolsep{3pt}
  \centering
  \caption{Effectiveness of each loss related to domain adaptation.}
  \renewcommand\arraystretch{1.2}
  \begin{tabular}{ccccc|cc}
    \Xhline{1pt}
    \multicolumn{1}{c}{$\mathcal{L}^{geo}_{flow}$}&
    \multicolumn{1}{c}{$\mathcal{L}^{consis}_{flow}$}&
    \multicolumn{1}{c}{$\mathcal{L}^{contra}_{warpErr}$}&    \multicolumn{1}{c}{$\mathcal{L}^{self}_{flow}$}&
    \multicolumn{1}{c|}{$\mathcal{L}^{kl}_{corr}$}&
    \multicolumn{1}{c}{EPE}& \multicolumn{1}{c}{Fl-all}\\

    \hline
    $\times$& $\times$& $\times$ & $\times$ & $\times$& 2.94 &30.85\% \\
    $\surd$& $\times$& $\times$ & $\times$ & $\times$& 2.87 &29.95\% \\
    $\times$& $\surd$& $\times$ & $\times$ & $\times$& 1.60 &14.12\% \\
    $\surd$& $\surd$& $\times$ & $\times$ & $\times$& 1.35 &11.20\% \\
    $\surd$& $\surd$& $\surd$ & $\times$ & $\times$& 1.20 &10.04\% \\
    $\surd$& $\surd$& $\surd$ & $\surd$ & $\times$& 1.09 &9.90\% \\
    $\surd$& $\surd$& $\surd$ & $\times$ & $\surd$& 0.88 &8.53\% \\
    $\surd$& $\surd$& $\surd$ & $\surd$ & $\surd$& \textbf{0.72} &\textbf{6.90\%} \\
       \Xhline{1pt}
  \end{tabular}
   \label{tab:Ablation_Loss}
\end{table}

\begin{table}
  \setlength\tabcolsep{3pt}
    \centering
    \caption{Ablation study on training data and flow backbones. The impact of training data sequence on optical flow surpasses various flow backbones.}
    \renewcommand\arraystretch{1.1}
    \begin{tabular}{c|c|c|c}
      \Xhline{1pt}
      \multicolumn{1}{c|}{Training data} & Method & EPE & Fl-all \\
      \hline
      \multicolumn{1}{c|}{\multirow{3}{*}{\makecell{\textbf{C}lean,\\\textbf{S}ynthetic degraded,\\\textbf{R}eal degraded}}}& Ours w/ RAFT & 0.75 & 7.01\% \\
      \cline{2-4}
				\multicolumn{1}{c|}{}& Ours w/ GMA & 0.74 & 6.96\% \\
      \cline{2-4}
				\multicolumn{1}{c|}{}& Ours w/ FlowFormer & \textbf{0.72} & \textbf{6.90\%} \\
      \hline
      \multicolumn{1}{c|}{\multirow{3}{*}{\makecell{\textbf{C}lean,\\\textbf{R}estored,\\\textbf{R}eal degraded}}}& Ours w/ RAFT & 1.02 & 9.13\% \\
      \cline{2-4}
				\multicolumn{1}{c|}{}& Ours w/ GMA & 0.97 & 8.60\% \\
      \cline{2-4}
				\multicolumn{1}{c|}{}& Ours w/ FlowFormer & 0.96 & 8.54\% \\
         \Xhline{1pt}
    \end{tabular}

    \label{tab:Ablation_training_data}
\end{table}

\begin{figure*}
  \centering
   \includegraphics[width=0.99\linewidth]{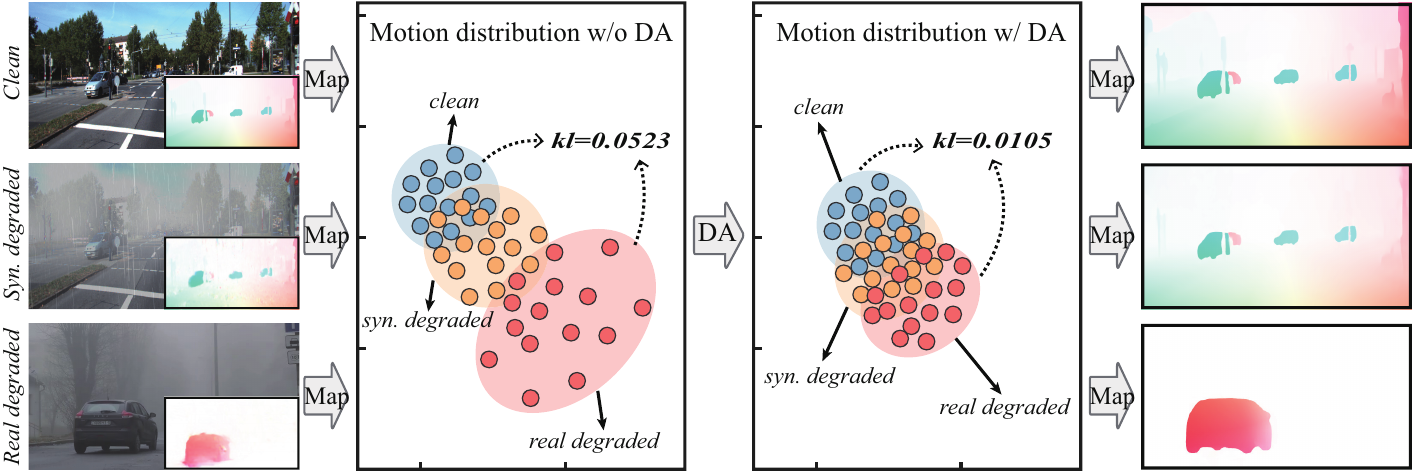}
   \caption{Optical flow visualization and t-SNE visualization of motion features. Without any domain adaptation, there exists a gap between clean domain and real degraded domain, while synthetic degraded domain serves as an intermediate bridge across the two domains. With cumulative domain adaptation, the three domains are further closer to each other, and the corresponding optical flows are improved well.
   }
   \label{Discussion_Bridge}

\end{figure*}

\subsection{Ablation Study}
\textbf{Effectiveness of Cumulative Adaptation Architecture.}
In Fig. \ref{Ablation_Cumulative}, we demonstrate the effectiveness of the proposed cumulative adaptation framework. Without any adaptation, namely directly estimating optical flow from degraded images using the pre-trained model, the optical flow model cannot work normally. With only clean-degraded motion adaptation, the optical flow performance of synthetic dataset is obviously improved, while there exists an upper limit in real dataset. \review{The main reason is that the large domain gap between clean and real degraded domains is caused by clean-degraded adverse weather and synthetic-real scene style, limiting the performance in real degraded scenes.} With synthetic-real motion adaptation, the optical flow performance shows only minor improvements in both synthetic and real datasets, indicating that clean-degraded gap has a dominant impact on optical flow, with the synthetic-real gap being secondary. When using the proposed cumulative adaptation framework for both clean-degraded and synthetic-real gaps, there are significant improvements in both synthetic and real datasets. Therefore, the proposed cumulative framework is crucial for adverse weather optical flow.

\begin{table}
    \setlength\tabcolsep{2pt}
  \centering
  \caption{Choice of different sampling strategies in contrastive learning, including random sampling, edge-aware sampling and entropy-aware sampling.}
  \renewcommand\arraystretch{1.1}
  \begin{tabular}{cc|cccc}
    \Xhline{1pt}
      \multicolumn{1}{c|}{\multirow{2}{*}{Clean domain}} & \multicolumn{1}{c|}{\multirow{2}{*}{Degraded domain}} &
      \multicolumn{2}{c|}{Rain-KITTI2015} & \multicolumn{2}{c}{Rain-GOF}\\
      \cline{3-6}
      \multicolumn{1}{c|}{} & \multicolumn{1}{c|}{} & \multicolumn{1}{c|}{EPE} &\multicolumn{1}{c|}{Fl-all} & \multicolumn{1}{c|}{EPE} & \multicolumn{1}{c}{Fl-all}\\
      \hline
      \multicolumn{1}{c|}{Random} & \multicolumn{1}{c|}{Random} & \multicolumn{1}{c|}{5.83} & \multicolumn{1}{c|}{35.40\%} & \multicolumn{1}{c|}{0.87} & \multicolumn{1}{c}{8.02\%} \\
      \hline
      \multicolumn{1}{c|}{Random} & \multicolumn{1}{c|}{Entropy-aware} & \multicolumn{1}{c|}{5.79} & \multicolumn{1}{c|}{34.86\%} & \multicolumn{1}{c|}{0.86} & \multicolumn{1}{c}{7.98\%} \\
      \hline
      \multicolumn{1}{c|}{Edge-aware} & \multicolumn{1}{c|}{Random} & \multicolumn{1}{c|}{5.42} & \multicolumn{1}{c|}{31.25\%} & \multicolumn{1}{c|}{0.76} & \multicolumn{1}{c}{7.11\%} \\
      \hline
      \multicolumn{1}{c|}{Edge-aware} & \multicolumn{1}{c|}{Entropy-aware} & \multicolumn{1}{c|}{\textbf{5.35}} & \multicolumn{1}{c|}{\textbf{30.29\%}} & \multicolumn{1}{c|}{\textbf{0.74}} & \multicolumn{1}{c}{\textbf{7.05\%}} \\

       \Xhline{1pt}
  \end{tabular}
   \label{tab:Discussion_Contrastive}
\end{table}

\noindent
\textbf{Role of Homogeneous and Heterogeneous Adaptation.}
We illustrate the impact of homogeneous and heterogeneous adaptation in the process of knowledge transfer in Table \ref{tab:Ablation_HomeHete}. \review{Note that the homogeneous adaptation here is to align the similar features including the intrinsic motion of the scene, clean boundary in warp error and correlation distribution in cost volume, and the heterogeneous adaptation here is to strip away the degraded motion features caused by adverse weather, \emph{e.g.}, degraded boundary errors.} Without knowledge transfer, degraded optical flow estimation is failed. With the cumulative framework based on implicit features, there is a significant improvement in optical flow performance, indicating that cumulative framework is the key to adverse weather optical flow. With the cumulative framework based on homogeneous adaptation, the optical flow metric is further improved. This is because the homogeneous adaptation can explicitly model the transfer process of intrinsic motion patterns. With only cumulative framework based on heterogeneous adaptation, the improvement in optical flow performance is not as pronounced as the homogeneous adaptation. When using homogeneous and heterogeneous adaptation techniques, optical flow performance is superior to the homogeneous adaptation alone. This reveals that homogeneous adaptation is crucial for ensuring the fundamental performance of adverse weather optical flow, while heterogeneous adaptation further optimizes the details of motion.

\begin{figure}
  \centering
   \includegraphics[width=0.99\linewidth]{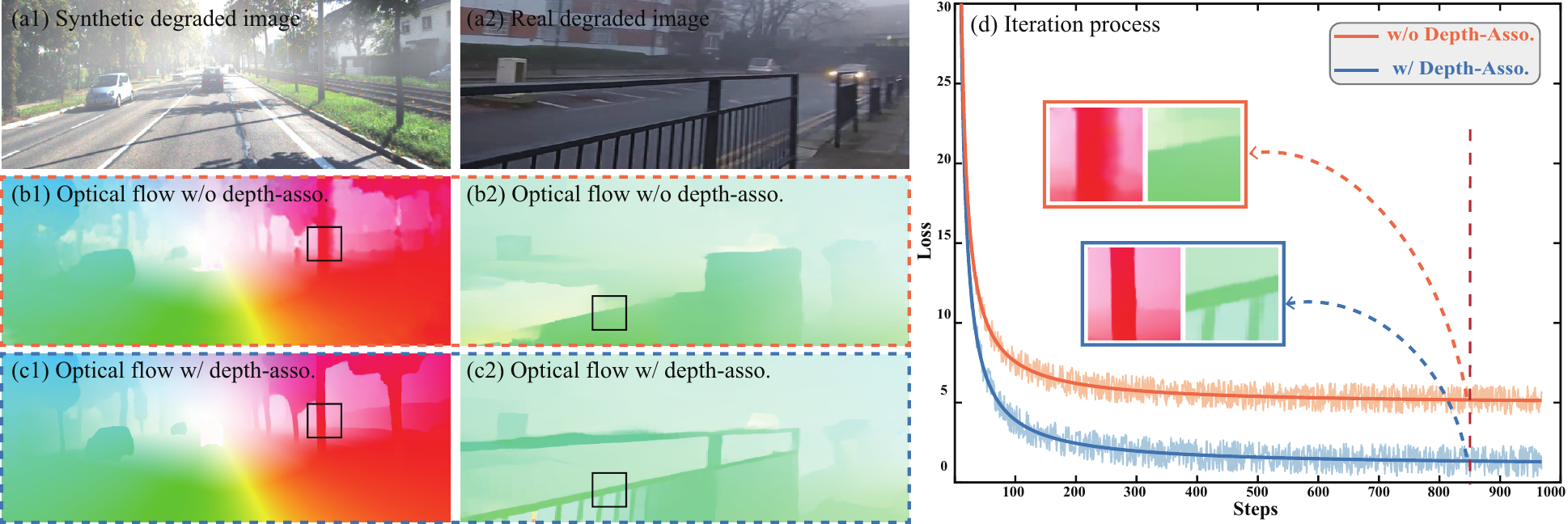}
   \caption{Effect of depth association transfer on optical flow. (a) Synthetic and real degraded images. Optical flows estimated (b) without depth association transfer and (c) with depth association transfer. (d) Iteration process. Depth association transfer can improve rigid motion boundary.
   }
   \label{Discussion_Depth}

\end{figure}

\noindent
\textbf{Effectiveness of Adaptation Losses.}
We study how the motion adaptation losses of the proposed method contribute to the final result as shown in Table \ref{tab:Ablation_Loss}. We set three experimental datasets, including synthetic Rain-KITTI2015, synthetic Fog-KITTI2015 and real Weather-GOF. We have four observations. First, $\mathcal{L}^{geo}_{flow}$ can improve the upper limit of transfer ability of intrinsic motion knowledge from clean domain to degraded domain.
Second, $\mathcal{L}^{consis}_{flow}$ makes a major contribution to the optical flow result of degraded domain. Third, $\mathcal{L}^{contra}_{warpErr}$ can further refine the optical flow of dynamic weather. Fourth, $\mathcal{L}^{self}_{flow}$ and $\mathcal{L}^{kl}_{corr}$ improve the optical flow performance in real degraded scenes.

\begin{figure*}
  \centering
   \includegraphics[width=1.0\linewidth]{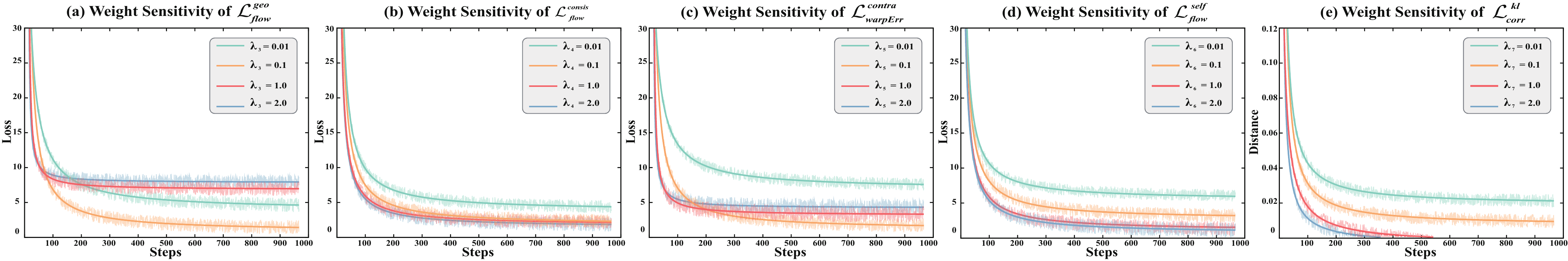}
   \caption{The weight sensitivity of each loss of motion adaptation during the training phase. The proposed framework is sensitive to the flow geometry loss, the contrastive loss of warp error and the KL divergence loss of correlation distribution, while is robust to other losses.
   }
   \label{Discussion_WeightSensitivity}

\end{figure*}

\noindent
\textbf{Influence of Training Data and Backbone.}
In Table \ref{tab:Ablation_training_data}, we compare the influence of various backbones (\emph{e.g.}, RAFT, GMA and FlowFormer) and training data settings (\emph{e.g.}, clean-synthetic degraded-real degraded (CSR) images, clean-restored-real degraded (CRR) images) on the final optical flow result. \review{Note that the restored image is the restoration result of the real degraded image.} As for the flow backbones, there is no obvious difference in the improvement of optical flow results from various flow backbones. As for the training data set, CSR knowledge transfer is superior to CRR. The main reason is that, the former can divide the gap between clean and real degraded domains into two sub-gaps for alleviating the feature distribution discrepancy, while in the latter, residual degradation in the restored image may exacerbate this gap. This shows that reasonable training data setting is more conducive to the proposed framework than the flow backbone to learn the intrinsic motion patterns of the scene.

\begin{table}
    \setlength\tabcolsep{1.5pt}
  \centering
  \caption{\review{Discussion on the impacts of different depth estimation approaches on the proposed training framework. ``mono.'' is short for ``monocular''. ``Params'' denotes trainable parameters, and ``Time'' denotes training time.}}
  \renewcommand\arraystretch{1.1}
  \begin{tabular}{c|cccccc}
    \Xhline{1pt}
    Method & Params (M) & Time (h) & EPE \\
    \hline
    Ours w/o depth & 110  & 30.2 & 1.18 \\

    Ours w/ mono. Marigold \cite{ke2024repurposing} & 1058  & 90.2 & 0.70 \\

    Ours w/ mono. DepthAnything \cite{yang2024depth} & 445 & 75.8 & 0.69 \\

    Ours w/ stereo AANet \cite{xu2020aanet} & 116 & 35.7 & 0.72 \\

       \Xhline{1pt}
  \end{tabular}
   \label{tab:depth_importance}
\end{table}

\subsection{Analysis and Discussion}
\textbf{Why does Intermediate Domain Play the Bridge?}
In Fig. \ref{Discussion_Bridge}, in order to illustrate the importance of synthetic degraded domain as the bridge between clean and real degraded domains, we visualize the motion feature distributions of different domains via t-SNE. Without any knowledge transfer, there is a large gap in the distribution of motion features between clean domain and real degraded domain where the features of real degraded domain are scattered, while synthetic degraded domain cross the clean and synthetic degraded domains to bridge this gap. With knowledge transfer, synthetic degraded domain pulls the distributions of the two domains together, and brings the motion features of real degraded domain closer to each other. Therefore, synthetic degraded domain facilitates a more directional knowledge transfer process.

\noindent
\textbf{\review{Effect of Depth Association on Flow Model Training?}}
\review{We study the effect of depth association in the training process of the optical flow model in Fig. \ref{Discussion_Depth}.} Without depth association in Fig. \ref{Discussion_Depth} (b), the rigid motion boundaries are blurry. With depth association in Fig. \ref{Discussion_Depth} (c), the optical flow is global-smooth with sharp boundaries. Besides, we visualize their training iteration process in Fig. \ref{Discussion_Depth} (d). \review{We can observe that depth association can further improve the optimal value that the optical flow model converges to.} Therefore, the depth association benefits enhancing the rigid motion boundary, thus guiding the intrinsic scene motion knowledge transfer from the clean domain to the degraded domain.

\begin{figure}
  \centering
   \includegraphics[width=0.99\linewidth]{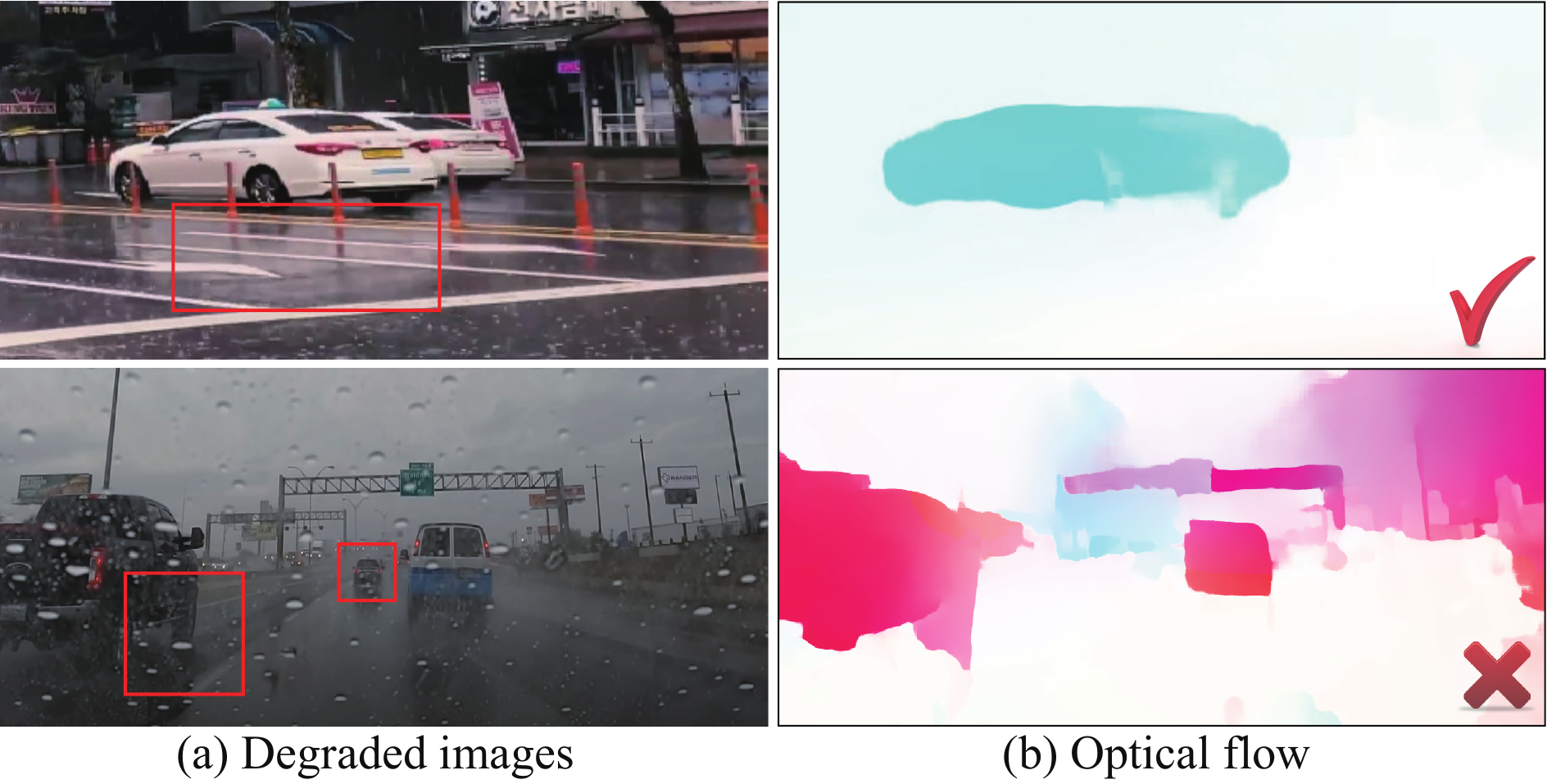}
   \caption{\review{Visualization of optical flow under various visual effects caused by rain. The proposed method is robust to the reflection from water on the ground (seeing the first row), while fails for the occlusion from raindrops falling on the camera (seeing the second row).}}
   \label{Fig_Robust_rainy}

\end{figure}

\noindent
\textbf{Impact of Contrastive Learning Sampling.}
The positive and negative sampling strategy is of great importance to the quality of the discriminative feature in contrastive learning. In Table \ref{tab:Discussion_Contrastive}, we show the advantage of the proposed edge-aware and entropy-aware sampling strategies over the conventional random sampling. Compared with random sampling, the edge-aware sampling strategy in the clean domain could greatly improve the optical flow. The main reason is most random sample patches in the warp error map are the meaningless regions with zero values. On the contrary, the sharp edge-aware sampling would guarantee us abundant discriminative information so as to better reflect the alignment of the optical flow boundary. Similarly, the entropy-aware sampling could also select the informative highly-informative boundary of warp error caused by degradations, facilitating to differ the flow boundary of the image structure from that of degradations.

\noindent
\textbf{Weight Sensitivity of Model Losses.}
To choose the optimal weight parameters for the total loss, we conduct analysis experiments on the weight sensitivity of the typical model losses, such as, $\mathcal{L}^{geo}_{flow}$, $\mathcal{L}^{consis}_{flow}$, $\mathcal{L}^{contra}_{warpErr}$, $\mathcal{L}^{self}_{flow}$ and $\mathcal{L}^{kl}_{corr}$. In Fig. \ref{Discussion_WeightSensitivity} (a), the larger the weight of $\mathcal{L}^{geo}_{flow}$, the more the depth dominates the network training, resulting in that optical flow may focus too much on rigid regions, but ignore the foreground non-rigid moving objects.
In Fig. \ref{Discussion_WeightSensitivity} (b), when the weight of the flow consistency loss $\mathcal{L}^{consis}_{flow}$ is greater than 1, the convergence speed of optical flow network reaches the bottleneck. In Fig. \ref{Discussion_WeightSensitivity} (c), a larger weight of the loss $\mathcal{L}^{contra}_{warpErr}$ instead contributes negatively to the final optical flow results. The main reason is that a too large weight may make the network pay too much attention to contrastive learning and ignore other discriminative features. In Fig. \ref{Discussion_WeightSensitivity} (d), the weight of $\mathcal{L}^{self}_{flow}$ can speed up the knowledge transfer from synthetic degraded domain to real degraded domain. When the weight is greater than 1, the convergence speed of optical flow network training is unchanged. In Fig. \ref{Discussion_WeightSensitivity} (e), we can observe that the correlation distribution alignment loss $\mathcal{L}^{kl}_{corr}$ is sensitive to the framework training. If the weight is too large, the gradient will disappear. Therefore, we set the adaptation losses weights $\{\lambda_3, \lambda_4, \lambda_5, \lambda_6, \lambda_7\}$ = $\{0.1, 1.0, 0.1, 1.0, 0.1\}$.

\noindent
\textbf{\review{How to Choose Depth Estimation Approach.}}
\review{Since depth association is crucial for the training of the proposed framework, the accuracy of depth naturally determines the performance of the optical flow. To explore how to choose the reasonable depth estimation approach, we compare the impacts of no depth, monocular-based depth (\emph{e.g.}, Marigold \cite{ke2024repurposing} based on diffusion model \cite{ho2020denoising} and DepthAnything \cite{yang2024depth} based on foundation model \cite{bommasani2021opportunities}),
and stereo-based depth (\emph{e.g.}, AANet \cite{xu2020aanet}) approaches on the proposed training framework, as shown in Table \ref{tab:depth_importance}. Note that we choose KITTI 2015 dataset as clean domain and Fog-GOF dataset as real degraded domain. We have two observations. First, using depth significantly improves the optical flow performance. Second, monocular-based depth estimation approaches Marigold and DepthAnything does show a slight improvement (\emph{e.g.}, EPE) in the optical flow performance over stereo-based depth estimation approach AAnet, but they also introduce more training costs (\emph{e.g.}, trainable parameters and training time). Therefore, regarding the choice of depth estimation approaches, we should consider their impacts on the proposed training framework, and choose a balance between efficiency and accuracy.}

\noindent
\textbf{\review{Robustness to Various Visual Effects Caused by Rain.}} \review{The proposed method could effectively address the degradation of transmission media on optical flow in adverse weather conditions. In Fig. \ref{Fig_Robust_rainy}, we discuss the robustness of the proposed method to other visual effects in rainy scenes, such as water on the ground and raindrops falling on the camera. We can observe that the proposed method is robust to the reflection from water on the ground, but fails for raindrops falling on the camera. This is because raindrops that remain relatively stationary with the camera for a short time cause the patch-level occlusion effect in the imaging, leading to the failure of optical flow estimation in the occlusion. Actually, since the raindrops are attached to the sensor, we can use a wiper to physically remove the visual effect caused by raindrops on the camera, thus mitigating their impact on optical flow.}

\begin{figure}
  \centering
   \includegraphics[width=0.99\linewidth]{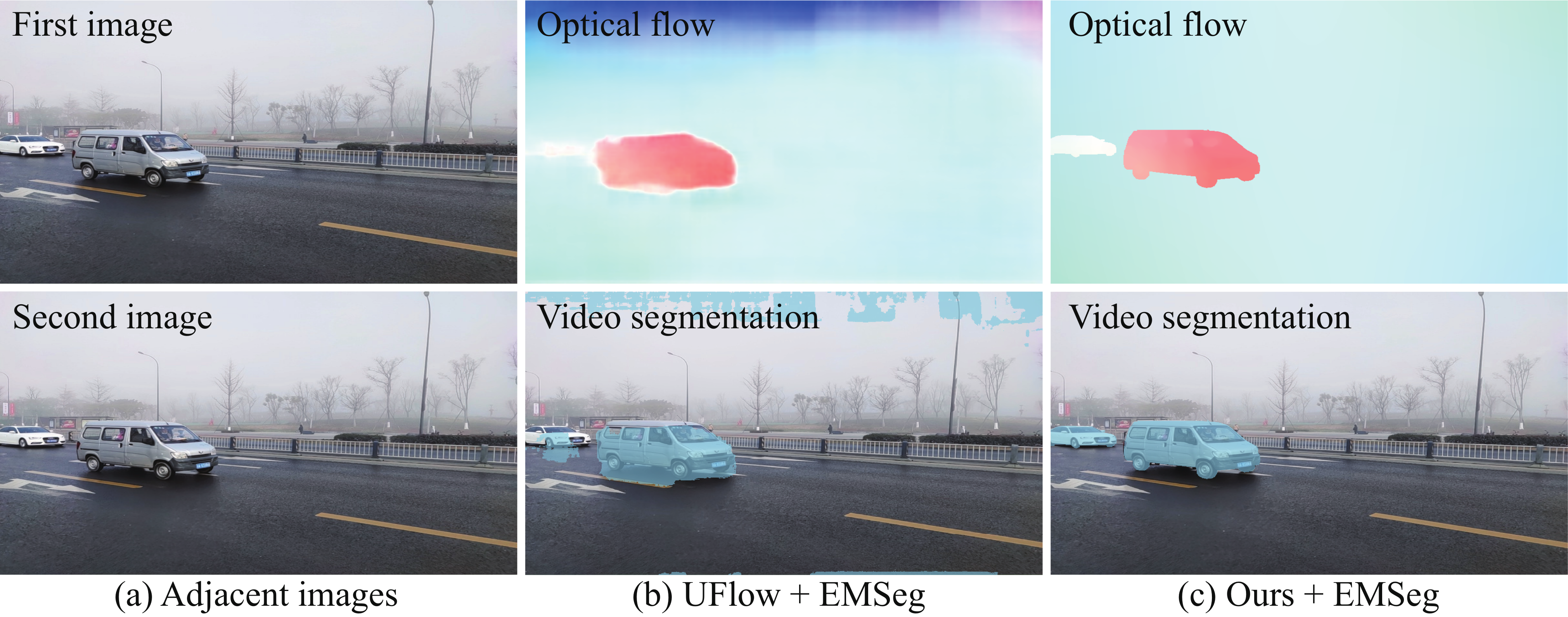}
   \caption{Visualization of video motion segmentation under adverse weather. The proposed method could not only improve the optical flow but also promote the interiority of video motion segmentation.
   }
   \label{Discussion_Donwstream}

\end{figure}

\begin{table}
  \setlength\tabcolsep{1pt}
    \centering
    \caption{Discussion on the model size and inference time on image 800 $\times$ 600.}
    \renewcommand\arraystretch{1.1}
    \begin{tabular}{c|c|c|c|c}
      \Xhline{1pt}
      Method & Runtime (ms) & Model size (M) & EPE & Fl-all \\
      \hline
      RobustFlow \cite{li2018robust} & 1843 & --& 12.25 & 80.93\% \\
      \hline
      DenseFogFlow \cite{yan2020optical} & 145 & 72.5 & 1.78 & 16.41\% \\
      \hline
      GyroFlow \cite{li2021gyroflow} & 55 & 24.6 & 0.95 & 9.13\% \\
      \hline
      UCDA-Flow \cite{zhou2023unsupervised} & 67 & 32.5 & 0.81 & 7.18\% \\
      \hline
      CH$^2$DA-Flow & 67 & 32.5 & 0.72 &  6.85\% \\
         \Xhline{1pt}
    \end{tabular}

    \label{tab:Discussion_Runtime}
\end{table}

\noindent
\textbf{Promotion for Downstream Video segmentation.}
Optical flow is a task of modeling the temporal correspondence between adjacent frames. To illustrate the promotion of optical flow for downstream vision task under adverse weather, we take video segmentation as an example, and compare the segmentation results of the same video motion segmentation method (\emph{e.g.}, EMSeg \cite{meunier2022driven}) with the optical flows directly estimated by UFlow \cite{jonschkowski2020matters} and the proposed method on degraded adjacent images.  When using the optical flow directly estimated by UFlow, the segmentation of independent moving objects is incomplete, and degraded background is included in the segmentation result. The main reason is that, degradation breaks the basic assumption of optical flow, thus leading to erroneous motion information provided to video segmentation. When using the optical flow from the proposed knowledge transfer framework, the segmentation result only include the independent moving object with complete structure, indicating that the proposed knowledge transfer method could effectively learn the intrinsic motion of the scene, thus promoting the downstream video segmentation.

\noindent
\textbf{Inference Time and Model Size.}
In Table \ref{tab:Discussion_Runtime}, on image size 800$\times$600, we compare the inference time and model size of our proposed method CH$^2$DA-Flow with those of other state-of-the-art methods, including optimization-based RobustFlow, semi-supervised DenseFogFlow, weakly-supervised GyroFlow and the previous version UCDA-Flow. \review{Note that the final optical flow model of the proposed method only needs the flow encoder and decoder of the real degraded domain for inference in Fig. \ref{Framework}.} The optimization-based method RobustFlow is time-consuming. The learning-based methods can reduce inference time. GyroFlow has the shortest inference time and smallest model size compared with DenseFogFlow and the proposed methods of the two versions. The main reason is that GyroFlow can directly obtain the background motion of optical flow from gyroscope data, significantly saving the inference time. On the contrary, although the inference speed of the proposed method is not the fastest and the simplest, the performance outperforms other methods by a significant margin. In addition, the previous version UCDA-Flow and the improved version CH$^2$DA-Flow have the same inference time and model size, but CH$^2$DA-Flow enhances the adverse weather optical flow performance. This indicates that the proposed homogeneous-heterogeneous adaptation is more suitable for adverse weather optical flow.

\begin{figure}
  \centering
   \includegraphics[width=0.99\linewidth]{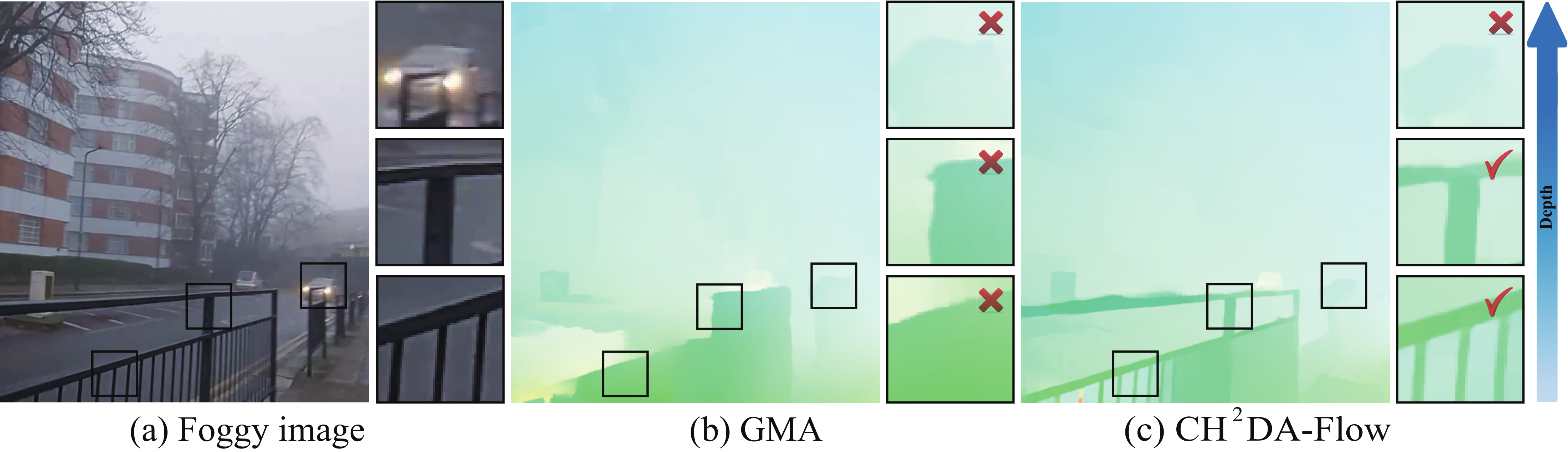}
   \caption{\review{Limitation of the proposed method. Compared with the state-of-the-art optical flow method GMA \cite{jiang2021transformer}, the proposed method obtains the clear motion boundary within a certain depth but fails for the too-distant moving objects under dense foggy conditions.}}
   \label{Fig_Limitation}

\end{figure}

\noindent
\textbf{\review{Limitation.}} \review{The proposed method can effectively achieve the optical flow estimation under most adverse weather conditions, but it is limited in dense foggy scenes, as shown in Fig. \ref{Fig_Limitation}. Compared with the state-of-the-art optical flow method GMA \cite{jiang2021transformer}, the proposed method obtains a clear motion boundary within a certain depth but fails for objects that are too distant under dense foggy conditions. There are two reasons for this problem. First, it is difficult for the stereo strategy of our framework to obtain accurate depth in distant regions. Second, fog is a non-uniform degradation related to depth, where the farther the distance, the more severe the degradation. This can lead to the loss of the too-distant moving objects, thus deteriorating the performance of optical flow. In the future, we will attempt to employ distance sensors (\emph{e.g.}, LiDAR and Radar) for detecting distant objects.}

\section{Conclusion}
In this work, we propose a novel cumulative homogeneous-heterogeneous domain adaptation framework for adverse weather optical flow, \review{which introduces synthetic degraded domain as an intermediate bridge between clean domain and real degraded domain to learn the cross-domain discriminative feature representations.} To the best of our knowledge, we are the first to investigate the cumulative transfer learning to tackle the problem of the large inter-domain gap and introduce the homogeneous-heterogeneous adaptation technique to explicitly model the process of knowledge transfer for optical flow estimation task under adverse weather. Within the unified cumulative framework, we propose depth association to consistently transfer the homogeneous intrinsic scene motion from clean scene to static weather, and introduce warp error to contrastively strip away the heterogeneous motion boundary errors caused by dynamic weather, and then build cost volume to holistically distill the homogeneous correlation distribution from synthetic to real adverse weather. In addition, we propose a real adverse weather dataset with optical flow labels, and conduct extensive experiments on public datasets and the proposed dataset to demonstrate that the proposed method significantly outperforms the state-of-the-art methods. We believe our work could not only facilitate the development of the adverse weather optical flow but also enlighten the researchers of the broader filed, \emph{i.e.}, scene understanding under adverse weather.

\bibliographystyle{IEEEtran}
\bibliography{egbib}

\begin{IEEEbiography}[{\includegraphics[width=1in, height=1.30in, clip, keepaspectratio]{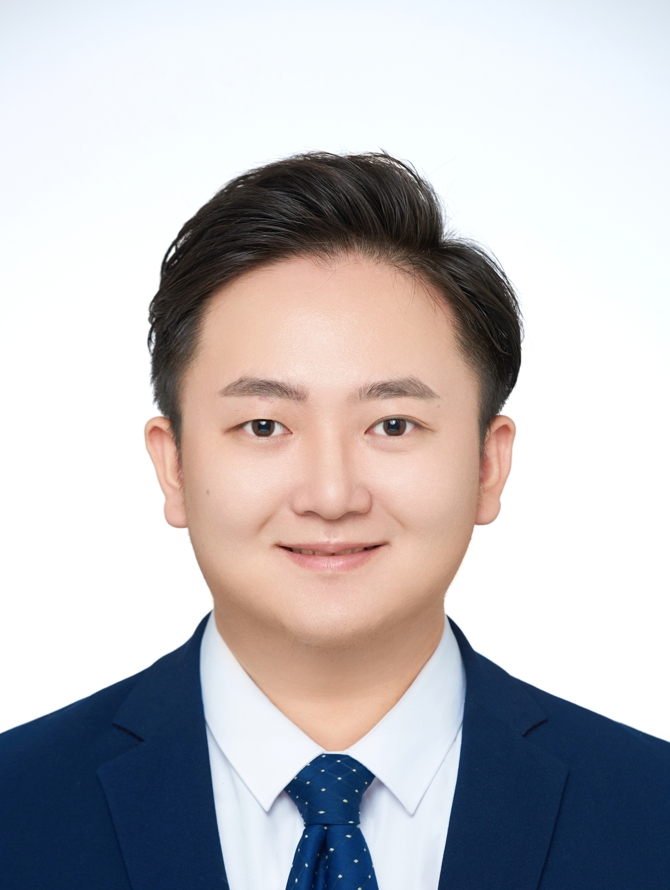}}]{Hanyu Zhou} received the B.E. degree in optoelectronic information science and engineering from Central South University, Changsha, China, in 2019. He is currently pursuing Ph.D. degree with the School of Artificial Intelligence and Automation, Huazhong University of Science and Technology, Wuhan, China. His research interests include motion estimation under adverse conditions and domain adaptation.
\end{IEEEbiography}

\begin{IEEEbiography}[{\includegraphics[width=1in, height=1.30in, clip, keepaspectratio]{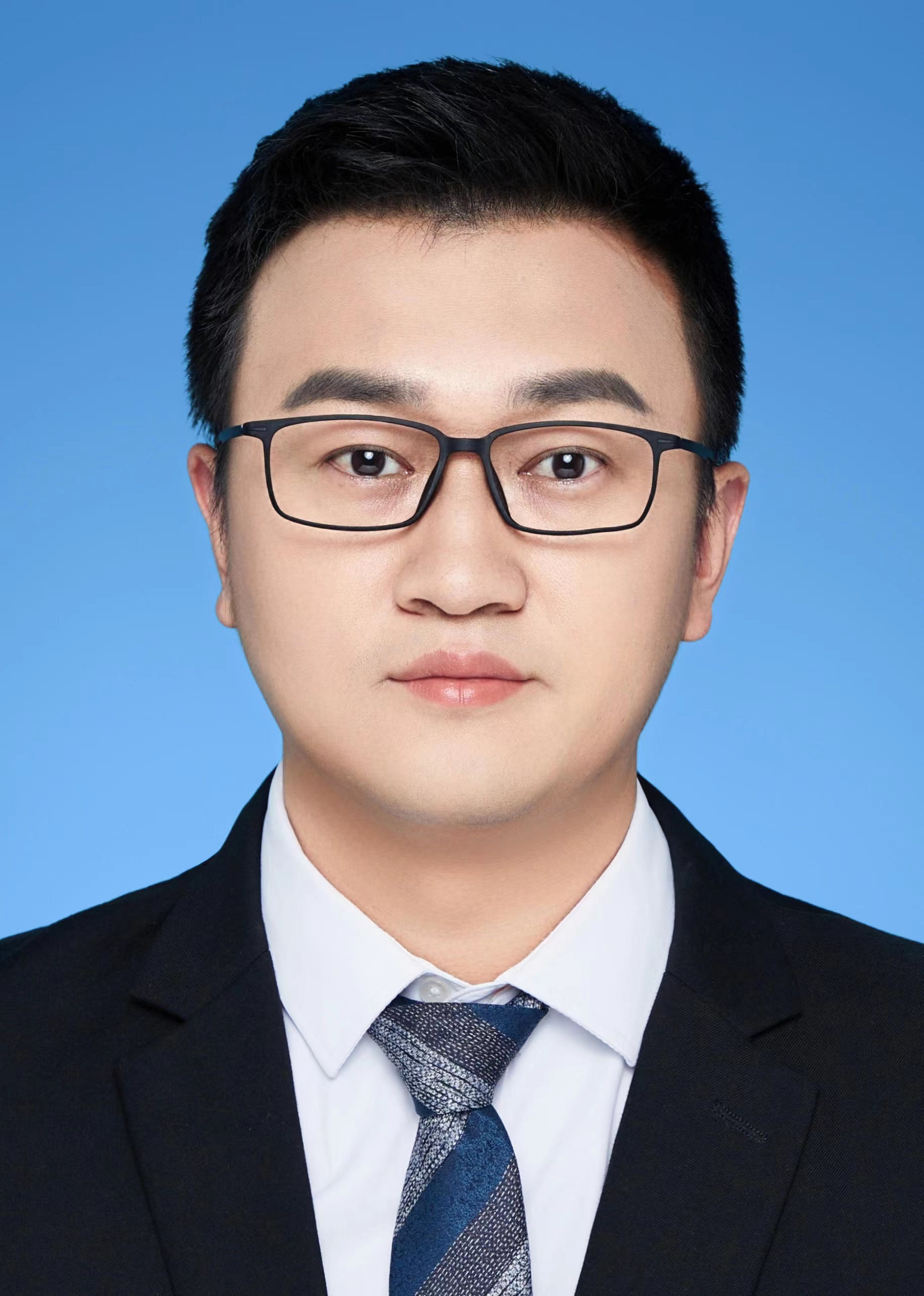}}]{Yi Chang} received B.S. degree from University of Electronic Science and Technology of China, Chengdu, China, in 2011, and M.S. degree and Ph.D. degree from Huazhong University of Science and Technology (HUST), in 2014 and 2019, respectively. He is currently an Associate Professor with School of Artificial Intelligence and Automation, Huazhong University of Science and Technology. His research interests include image restoration and understanding.
\end{IEEEbiography}

\begin{IEEEbiography}[{\includegraphics[width=1in, height=1.30in, clip, keepaspectratio]{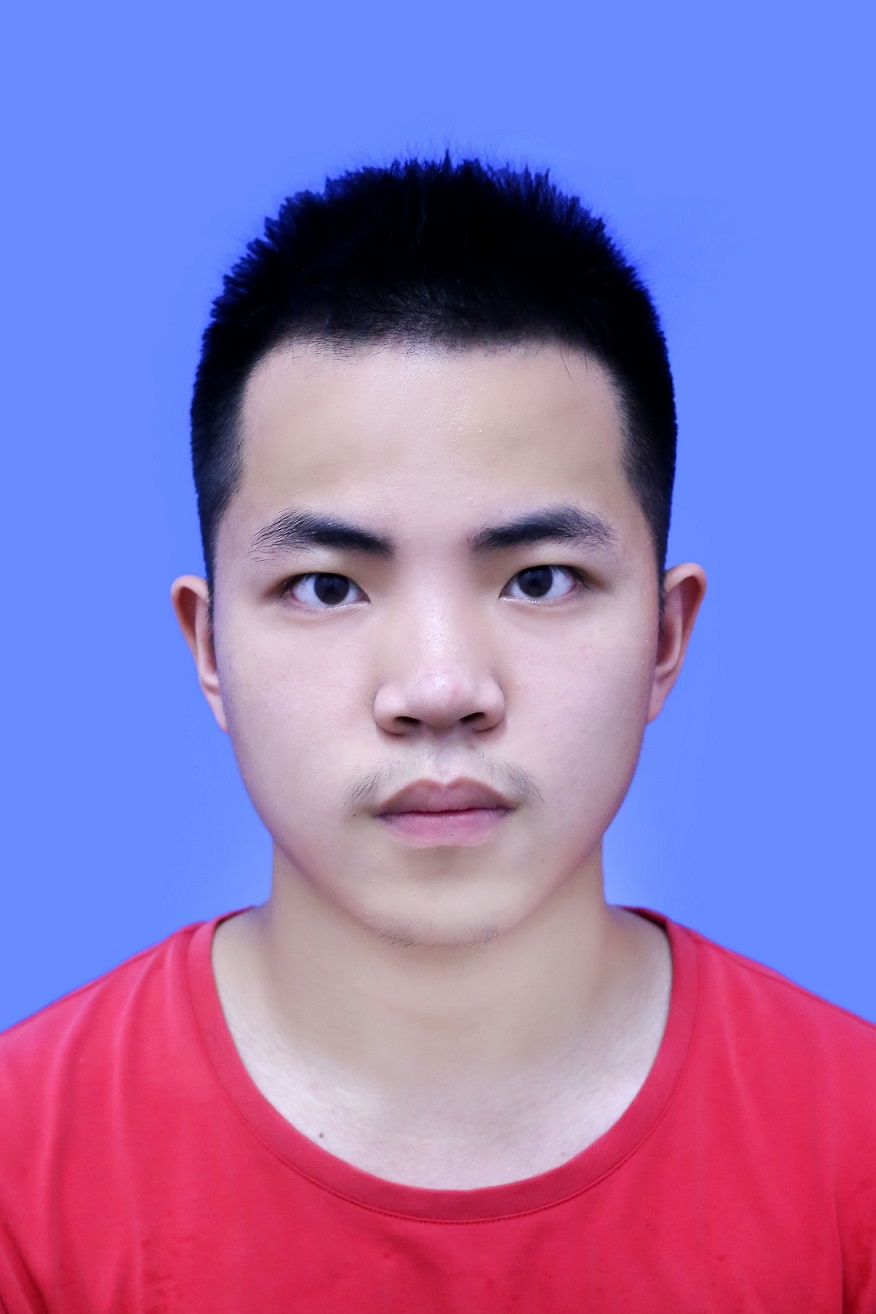}}]{Zhiwei Shi} received the B.E. degree in electronic science and technology from Huazhong University of Science and Technology, Wuhan, China, in 2023. He is currently pursuing M.S. degree with the School of Artificial Intelligence and Automation, Huazhong University of Science and Technology, Wuhan, China. His research interests include motion segmentation and multimodal fusion.
\end{IEEEbiography}

\begin{IEEEbiography}[{\includegraphics[width=1in, height=1.30in, clip, keepaspectratio]{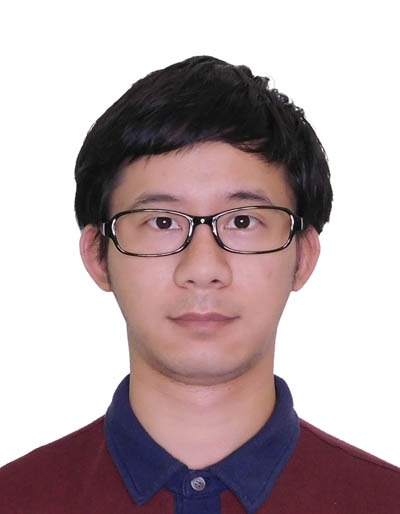}}]{Wending Yan} received B.E. degree in engineering, electrical and electronic engineering from Nanyang Technological University, in 2016, and Ph.D. degree in electrical and computer engineering from National University of Singapore, Singapore, in 2022, respectively. He is currently a Senior Engineer with Huawei International Co. Ltd., Singapore. His interests include adverse weather motion estimation and image restoration.
\end{IEEEbiography}

\begin{IEEEbiography}[{\includegraphics[width=1in, height=1.30in, clip, keepaspectratio]{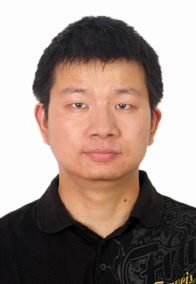}}]{Gang Chen} received B.E. degree in biomedical engineering and B.S. degree in mathematics and applied mathematics, and M.S. degree in control science and engineering from Xi'an Jiaotong University, China, in 2008 and 2011, respectively, and Ph.D. degree from Technical University of Munich, Germany, in 2016. He is currently a Professor with Sun Yat-sen University, Guangzhou, China. His research interests include high-performance computing and embedded systems.
\end{IEEEbiography}

\begin{IEEEbiography}[{\includegraphics[width=1in, height=1.30in, clip, keepaspectratio]{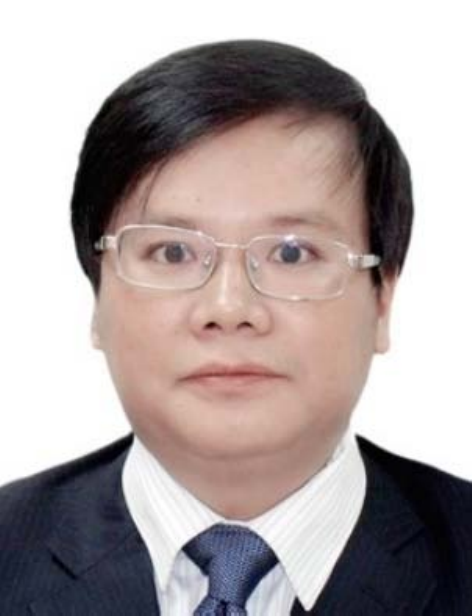}}]{Yonghong Tian} is currently a Boya Distinguished Professor with the Department of Computer Science and Technology, Peking University, China, and is also the deputy director of Artificial Intelligence Research Center, PengCheng Laboratory, Shenzhen, China. His research interests include neuromorphic vision and multimedia big data. He has published over 200 technical articles in IEEE TPAMI/NeurIPS/CVPR. Prof. Tian was/is an Associate Editor of IEEE TCSVT, IEEE TMM, IEEE Multimedia Magazine.
\end{IEEEbiography}

\begin{IEEEbiography}[{\includegraphics[width=1in, height=1.30in, clip, keepaspectratio]{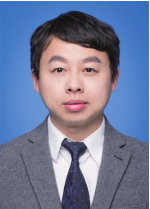}}]{Luxin Yan} received the B.S. degree in electronic communication engineering, and the Ph.D. degree in pattern recognition and intelligence system from Huazhong University of Science and Technology (HUST) in 2001 and 2007, respectively. He is currently a Professor with School of Artificial Intelligence and Automation, HUST. His research interests include multi-spectral image processing, pattern recognition and real-time embedded system.
\end{IEEEbiography}

\end{document}